%% file: cl2017.tex
\documentclass{clv3}
\historydates{Submission received: 30 June 2018; revised version received: 20 March 2019; accepted for publication: 12 June 2019.}

\runningtitle{Modeling Language Variation and Universals}

\runningauthor{Ponti et al.}

\usepackage{latexsym}
\usepackage{amsmath}
\usepackage{booktabs}
\usepackage{tabularx}
\usepackage{multirow}
\usepackage{microtype}
\usepackage[utf8]{inputenc}
\usepackage[T1]{fontenc}
\usepackage{graphicx}
\usepackage{url}
\usepackage{tipa}
\usepackage{algorithm}
\usepackage[noend]{algpseudocode}
\usepackage{caption}
\usepackage{subcaption}
\usepackage{ragged2e}

\newcolumntype{R}{>{\raggedright\arraybackslash}X}

\newcolumntype{b}{X}
\newcolumntype{m}{>{\hsize=.43\hsize}R}
\newcolumntype{s}{>{\hsize=.2\hsize}R}
\newcolumntype{n}{>{\hsize=.35\hsize}R}

\usepackage{hyperref}
\usepackage{cleveref}
\usepackage[table,x11names,dvipsnames,table]{xcolor}
\definecolor{darkblue}{rgb}{0, 0, 0.5}
\hypersetup{colorlinks=true,citecolor=darkblue, linkcolor=darkblue, urlcolor=darkblue}

\definecolor{LighterGray}{gray}{0.92}

\usepackage{tikz}
\usepackage{tikz-qtree}
\usetikzlibrary{arrows}
\usepackage{todonotes}

\begin{document}

\title{Modeling Language Variation and Universals:
A Survey on Typological Linguistics for
Natural Language Processing}

\author{Edoardo Maria Ponti\thanks{English Faculty Building, 9 West Road 
Cambridge CB3 9DA, United Kingdom. E-mail: ep490@cam.ac.uk}}
\affil{LTL, University of Cambridge}
\author{Helen O'Horan\thanks{English Faculty Building. E-mail: helen.ohoran@gmail.com}}
\affil{LTL, University of Cambridge}
\author{Yevgeni Berzak\thanks{77 Massachusetts Avenue, Cambridge, MA 02139, USA. E-mail: berzak@mit.edu}}
\affil{Department of Brain and Cognitive Sciences, MIT}
\author{Ivan Vuli\'{c}\thanks{English Faculty Building. E-mail: iv250@cam.ac.uk}}
\affil{LTL, University of Cambridge}
\author{Roi Reichart\thanks{Technion City, Haifa 3200003, Israel. E-mail: roiri@ie.technion.ac.il}}
\affil{Faculty of Industrial Engineering and Management, Technion -- IIT}
\author{Thierry Poibeau\thanks{1 Rue Maurice Arnoux, 92120 Montrouge, France. E-mail: thierry.poibeau@ens.fr}}
\affil{LATTICE Lab, CNRS and ENS/PSL and Univ. Sorbonne nouvelle/USPC}
\author{Ekaterina Shutova\thanks{Science Park 107, 1098 XG Amsterdam, Netherlands. E-mail: shutova.e@uva.nl}}
\affil{ILLC, University of Amsterdam}
\author{Anna Korhonen\thanks{English Faculty Building. E-mail: alk23@cam.ac.uk}}
\affil{LTL, University of Cambridge}

\maketitle

\begin{abstract}
Linguistic typology aims to capture structural and semantic variation across the world’s languages. A large-scale typology could provide excellent guidance for multilingual Natural Language Processing (NLP), particularly for languages that suffer from the lack of human labeled resources. We present an extensive literature survey on the use of typological information in the development of NLP techniques. Our survey demonstrates that to date, the use of information in existing typological databases has resulted in consistent but modest improvements in system performance. We show that this is due to both intrinsic limitations of databases (in terms of coverage and feature granularity) and under-employment of the typological features included in them. We advocate for a new approach that adapts the broad and discrete nature of typological categories to the contextual and continuous nature of machine learning algorithms used in contemporary NLP. In particular, we suggest that such approach could be facilitated by recent developments in data-driven induction of typological knowledge.

\end{abstract}

\section{Introduction}
\input{1_intro}

\section{Overview of Linguistic Typology}
\label{sec:overview}
\input{2_typology}

\section{Overview of Multilingual NLP}
\label{sec:applications}
\input{3_polyglotnlp}

\section{Selection and Development of Typological Information}
\label{sec:survey-development}
\input{4_development}

\section{Uses of Typological Information in NLP Models}
\label{sec:survey-applications}
\input{5_use}

\section{Future Research Avenues}
\label{sec:future}
\input{6_future}

\section{Conclusions}
\label{sec:commentary-conclusion}
\input{7_conclusions}

\section*{Acknowledgments}
This work is supported by ERC Consolidator Grant LEXICAL (no 648909). 

\starttwocolumn
\bibliography{cl2017}

\end{document}

%% file: 1_intro.tex
The world's languages may share universal features at a deep, abstract level, but the structures found in real-world, surface-level texts can vary significantly. This cross-lingual variation has challenged the development of robust, multilingually applicable NLP technology, and as a consequence, existing NLP is still largely limited to a handful of resource-rich languages. The architecture design, training, and hyper-parameter tuning of most current algorithms are far from being language-agnostic, and often inadvertently incorporate language-specific biases \cite{bender2009linguistically,Bender-2011}. In addition, most state-of-the-art machine learning models rely on supervision from (large amounts of) labeled data --- a requirement that cannot be met for the majority of the world's languages \citep{snyder-10}.

Over time, approaches have been developed to address the data bottleneck in multilingual NLP. These include unsupervised models that do not rely on the availability of manually annotated resources \citep[\textit{inter alia}]{snyder2008unsupervised,vulic2011identifying} and techniques that transfer data or models from resource-rich to resource-poor languages \cite[\textit{inter alia}]{Pado-2005,das-petrov-2011,Tackstrom-2012}. 
Some multilingual applications, such as Neural Machine Translation and Information Retrieval, have been facilitated by learning joint models that learn from several languages \cite[\textit{inter alia}]{Ammar-2016,johnson2016google} or via multilingual distributed representations of words and sentences \cite[\textit{inter alia}]{mikolov2013exploiting}. Such  techniques can lead to significant improvements in performance and parameter efficiency over monolingual baselines \citep{pappas2017multilingual}.

Another, highly promising source of information for modelling cross-lingual variation can be found in the field of Linguistic Typology. This discipline aims to systematically compare and document the world's languages based on the empirical observation of their variation with respect to cross-lingual benchmarks \cite{comrie1989language,croft2002typology}. Research efforts in this field have resulted in large typological databases, e.g. most prominently the World Atlas of Language Structures (WALS) \citep{wals-2013}. Such databases can serve as a source of guidance for feature choice, algorithm design, and data selection or generation in multilingual NLP.

Previous surveys on this topic have covered earlier research integrating typological knowledge into NLP \citep{o2016survey,bender2016linguistic}, however, there is still no consensus on the general effectiveness of this approach. For instance,   \citet{sproat2016language} has argued that data-driven machine learning should not need to commit to any assumptions about categorical and manually defined language types as defined in typological databases. 

In this paper, we provide an extensive survey of typologically informed NLP methods to date, including the more recent neural approaches not previously surveyed in this area. We consider the impact of typological (including both structural and semantic) information on system performance and discuss the optimal sources for such information. Traditionally, typological information has been obtained from hand-crafted databases and, therefore, it tends to be coarse-grained and incomplete. Recent research has focused on inferring typological information automatically from multilingual data \citep[\textit{inter alia}]{asgari2017past}, with the specific purpose of obtaining a more complete and finer-grained set of feature values. We survey these techniques and discuss ways to integrate their predictions into the current NLP algorithms. To the best of our knowledge, this has not yet been covered in the existing literature.

In short, the key questions our paper addresses can be summarized as follows: (i) Which NLP tasks and applications can benefit from typology? (ii) What are the advantages and limitations of currently available typological databases? Can data-driven inference of typological features offer an alternative source of information? (iii) Which methods have been proposed to incorporate typological information in NLP systems, and how should such information be encoded? (iv) To what extent does the performance of typology-savvy methods surpass typology-agnostic baselines? How does typology compare to other criteria of language classification, such as genealogy? (v) How can typology be harnessed for data selection, rule-based systems, and model interpretation?

We start this survey with a brief overview of Linguistic Typology (\autoref{sec:overview}) and multilingual NLP (\autoref{sec:applications}). After these introductory sections we proceed to examine the development of typological information for NLP, including that in hand-crafted typological databases and that derived through automatic inference from linguistic data (\autoref{sec:survey-development}).  In the same section, we also describe typological features commonly selected for application in NLP. In \autoref{sec:survey-applications} we discuss ways in which typological information has been integrated into NLP algorithms, identifying the main trends and comparing the performance of a range of methods. Finally, in \autoref{sec:future} we discuss the current limitations in the use of typology in NLP and propose novel research directions inspired by our findings. 

%% file: 2_typology.tex
There is no consensus on the precise number of languages in the world. For example, Glottolog provides the estimate of 7748 \cite{glottolog}, while Ethnologue \cite{ethnologue} refers to 7097.\footnote{These counts include only languages traditionally spoken by a community as their principal means of communication, and exclude unattested, pidgin, whistled, and sign languages.} This is due to the fact that defining what constitutes a 'language' is in part arbitrary. Mutual intelligibility, which is used as the main criterion for including different language variants under the same label, is gradient in nature. Moreover, social and political factors play a role in the definition of language.



Linguistic Typology is the discipline that studies the variation among the world's languages through their systematic comparison \cite{comrie1989language,croft2002typology}. The comparison is challenging because linguistic categories cannot be pre-defined \citep{haspelmath2007pre}. Rather, cross-linguistically significant categories emerge inductively from the comparison of known languages, and are progressively refined with the discovery of new languages. Crucially, the comparison needs to be based on functional criteria, rather than formal criteria. Typologists distinguish between \textit{constructions}, abstract and universal functions, and \textit{strategies}, the type of expressions adopted by each language to codify a specific construction \citep{croft2017linguistic}. For instance, the passive voice is considered a strategy that emphasizes the semantic role of patient: some languages lack this strategy and use others strategies to express the construction. For instance, Awtuw (Sepik family) simply allows for the subject to be omitted.

The classification of the strategies in each language is grounded in typological \textit{documentation} \cite[p. 248]{Bickel-2007}. Documentation is empirical in nature and involves collecting texts or speech excerpts, and assessing the features of a language based on their analysis. The resulting information is stored in large databases (see \S\ \ref{sec:overview-databases}) of attribute--values (this pair is henceforth referred to as \textit{typological feature}), where usually each attribute corresponds to a construction and each value to the most widespread strategy in a specific language.

\textit{Analysis} of cross-lingual patterns reveals that cross-lingual variation is bounded and far from random \cite{greenberg1966universals}. Indeed, typological features can be interdependent: the presence of one feature may implicate another (in one direction or both). This interdependence is called restricted universal, as opposed to unrestricted universals, which specify properties shared unconditionally by all languages. Such typological universals (restricted or not) are rarely absolute (i.e.\ exceptionless); rather, they are tendencies \cite{corbett2010implicational}, hence they are called ``statistical''. For example, consider this restricted universal: if a language (such as Hmong Njua, Hmong--Mien family) has prepositions, then genitive-like modifiers follow their head. If, instead, a language (such as Slavey, Na--Dené family) has postpositions, the order of heads and genitive-like modifiers is swapped. However, there are known exceptions: Norwegian (Indo--European) has prepositions but genitives precede their syntactic heads.\footnote{Exception-less generalizations are known as \textit{absolute universals}. However, properties that have been proposed as such are often controversial, because they are too vacuous or have been eventually falsified \citep{evans2009myth}.} Moreover, some typological features are rare while others are highly frequent. Interestingly, this also means that some languages are intuitively more plausible than others. Implications and frequencies of features are important as they unravel the deeper explanatory factors underlying the patterns of cross-linguistic variation \citep{dryer1998statistical}.
 
Cross-lingual variation can be found at all levels of linguistic structure. The seminal works on Linguistic Typology were concerned with morphosyntax, mainly morphological systems \cite[p. 128]{sapir2014language} and word order \cite{greenberg1966universals}. This level of analysis deals with the form of meaningful elements (morphemes and words) and their combination, hence it is called \textit{structural} typology. As an example, consider the alignment of the nominal case system \cite{dixon1994ergativity}: some languages like Nenets (Uralic) use the same case for subjects of both transitive and intransitive verbs, and a different one for objects (nominative--accusative alignment). Other languages like Lezgian (Northeast Caucasian) group together intransitive subjects and objects, and treat transitive subjects differently (ergative--absolutive alignment).


On the other hand, \textit{semantic} typology studies languages at the semantic and pragmatic levels. This area was pioneered by anthropologists interested in kinship \cite{d1995development} and colors \cite{berlin1969basic}, and was expanded by studies on lexical classes \citep{dixon1977have}. The main focus of semantic typology has been to categorize languages in terms of concepts \cite{Evans-2011} in the lexicon, in particular with respect to the 1) granularity, 2) division (boundary location), and 3) membership criteria (grouping and dissection). For instance, consider the event expressed by \textit{to open} (something). It lacks a precise equivalent in languages such as Korean, where similar verbs overlap in meaning only in part \cite{bowerman2001shaping}. For instance, \textit{ppaeda} means `to remove an object from tight fit', used e.g. for drawers, and \textit{pyeolchida} means `to spread out a flat thing', used e.g. for hands. Moreover, the English verb encodes the resulting state of the event, whereas an equivalent verb in another language such as Spanish (\textit{abrir}) rather expresses the manner of the event \cite{talmy1991path}. Although variation in the categories is pervasive due to their partly arbitrary nature, it is constrained cross-lingually via shared cognitive constraints \cite{majid2007semantic}.



Similarities between languages do not always arise from language-internal dynamics but also from external factors. In particular, similarities can be inherited from a common ancestor (genealogical bias) or borrowed by contact with a neighbor (areal bias) \citep{bakker2010language}. Owing to genealogical inheritance, there are features that are widespread within a family but extremely rare elsewhere (e.g. the presence of click phonemes in the Khoisan languages). As an example of geographic percolation, most languages in the Balkan area (Albanian, Bulgarian, Macedonian, Romanian, Torlakian) have developed, even without a common ancestor, a definite article that is put after its noun simply because of their close proximity.


Research in linguistic typology has sought to disentangle such factors and to integrate them into a single framework aimed at answering the question ``what's where why?'' \citep{nichols1992language}. Language can be viewed as a hybrid biological and cultural system. The two components co-evolved in a twin track, developing partly independently and partly via mutual interaction \citep{durham1991coevolution}. The causes of cross-lingual variation can therefore be studied from two complementary perspectives --- from the perspective of functional theories or event-based theories \citep{bickel2015distributional}. The former theories involve cognitive and communicative principles (internal factors) and account for the origin of variation, while the latter ones emphasize the imitation of patterns found in other languages (external factors) and account for the propagation (or extinction) of typological features \citep{croft1995autonomy,croft2000explaining}.


Examples of functional principles include factors associated with language use, such as the frequency or processing complexity of a pattern \citep{cristofaro1999ramat}. Patterns that are easy or widespread get integrated into the grammar \citep[\textit{inter alia}]{haspelmath1999optimality}. On the other hand, functional principles allow the speakers to draw similar inferences from similar contexts, leading to locally motivated pathways of diachronic change through the process known as \textit{grammaticalization} \citep{greenberg1966synchronic,greenberg1978diachrony,bybee1988diachronic}. For instance, in the world's languages (including English) the future tense marker almost always originates from verbs expressing direction, duty, will, or attempt because they imply a future situation.

The diachronic and gradual origin of the changes in language patterns and the statistical nature of the universals explain why languages do not behave monolithically. Each language can adopt several strategies for a given construction and partly inconsistent semantic categories. In other words, typological patterns tend to be gradient. For instance, the semantics of grammatical and lexical categories can be represented on continuous multi-dimensional maps \citep{croft2008inferring}. \citet{bybee2005alternatives} have noted how this gradience resembles the patterns learned by connectionist networks (and statistical machine learning algorithms in general). In particular, they argue that such architectures are sensitive to both local (contextual) information and general patterns, as well as to their frequency of use, similarly to natural languages.

Typological documentation is limited by the fact that the evidence available for each language is highly unbalanced and many languages are not even recorded in a written form.\footnote{According to \citet{ethnologue}, 34.4\% of the world's languages are threatened, not transmitted to younger generations, moribund, nearly extinct or dormant. Moreover, 34\% of the world's languages are vigorous but have not yet developed a system of writing.}
However, large typological databases such as WALS \citep{wals-2013} nevertheless have an impressive coverage (syntactic features for up to 1519 languages). Where such information can be usefully integrated in machine learning, it can provide an alternative form of guidance to manual construction of resources that are now largely lacking for low resource languages. We discuss the existing typological databases and the integration of their features into NLP models in sections 4 and 5.

%% file: 3_polyglotnlp.tex
The scarcity of data and resources in many languages represents a major challenge for multilingual NLP. Many state-of-the-art methods rely on supervised learning, hence their performance depends on the availability of manually crafted datasets annotated with linguistic information (e.g., treebanks, parallel corpora) and/or lexical databases (e.g., terminology databases, dictionaries). Although similar resources are available for key tasks in a few well-researched languages, the majority of the world's languages lack them almost entirely. This gap cannot be easily bridged: the creation of linguistic resources is a time-consuming process and requires skilled labor. Furthermore, the immense range of possible tasks and languages makes the aim of a complete coverage unrealistic.

One solution to this problem explored by the research community abandons the use of annotated resources altogether and instead focuses on unsupervised learning. This class of methods infers probabilistic models of the observations given some latent variables. In other words, it unravels the hidden structures within unlabeled text data. Although these methods have been employed extensively for multilingual applications \cite[\textit{inter alia}]{snyder2008unsupervised,vulic2011identifying,titov2012crosslingual}, their performance tends to lag behind the more linguistically informed supervised learning approaches \cite{Tackstrom-2013}. Moreover, they have been rarely combined with typological knowledge. For these reasons, we do not review them in this chapter.


Other promising ways to overcome data scarcity include transferring models or data from resource-rich to resource-poor languages (\autoref{ssec:langtrasf}) or learning joint models from annotated examples in multiple languages (\autoref{ssec:jointlearn}) in order to leverage language inter-dependencies. Early approaches of this kind have relied on universal, high-level delexicalized features, such as PoS tags and dependency relations. More recently, however, the incompatibility of (language-specific) lexica has been countered by mapping equivalent words into the same multilingual semantic space through representation learning (\autoref{subsec:reprlearn}). This has enriched language transfer and multilingual joint modelling with lexicalized features. In this chapter, we provide an overview of these methods, as they constitute the backbone of the typology-savvy algorithms surveyed in \autoref{sec:survey-applications}.

\subsection{Language Transfer}
\label{ssec:langtrasf}

\begin{figure}[t]
\centering
\includegraphics[width=\textwidth]{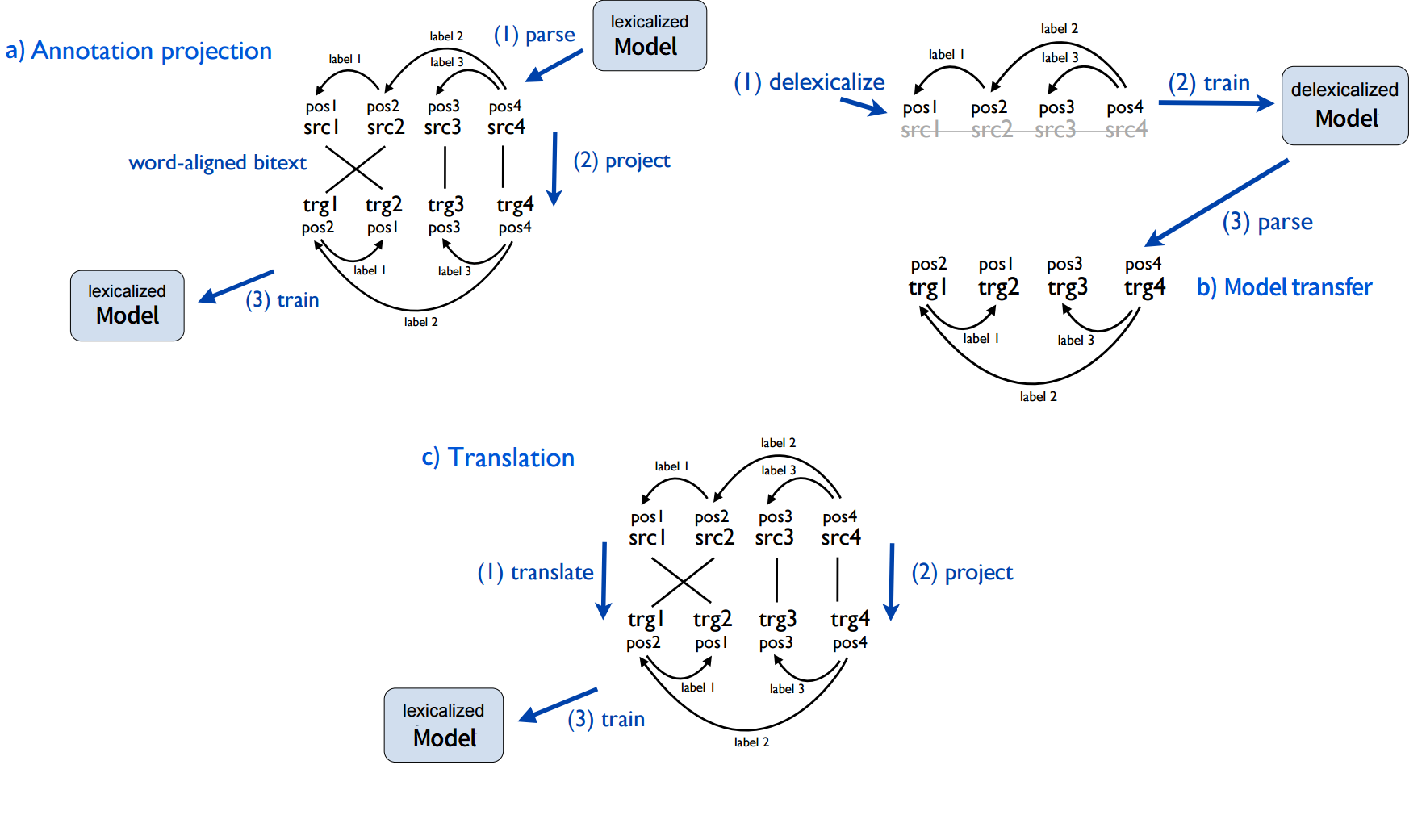}
\caption{Three methods for language transfer: a) annotation projection, b) model transfer, and c) translation. The image has been adapted from \protect\namecite{tiedemann2015cross}.}
\label{fig:langtransf}
\end{figure}

Linguistic information can be transferred from resource-rich languages to resource-poor languages: these are commonly referred to as source languages and target languages, respectively. Language transfer is challenging as it requires us to match word sequences with different lexica and word orders, or syntactic trees with different (anisomorphic) structures \citep{ponti2018isomorphic}. As a consequence, the information obtained from the source languages typically needs to be adapted, by tailoring it to the properties of the target languages. The methods developed for language transfer include annotation projection, (de)lexicalized model transfer, and translation \citep{agic2014cross}. We illustrate them below using dependency parsing as an example.

\textbf{Annotation projection} was introduced in the seminal work of \namecite{yarowsky2001inducing} and \citet{hwa-2005}. In its original formulation, as illustrated in Figure~\ref{fig:langtransf}a, a source text is parsed and word-aligned with a target parallel raw text. Its annotation (e.g. PoS tags and dependency trees) is then projected \textit{directly} and used to train a supervised model on the target language. Later refinements to this process are known as \textit{soft} projection, where constraints can be used to complement alignment, based on distributional similarity \cite{das-petrov-2011} or constituent membership \cite{pado2009cross}. Moreover, source model expectations on labels \citep{wang2014cross,agic2016multilingual} or sets of most likely labels \citep{Khapra-2011,Wisniewski:2014emnlp} can be projected instead of single categorical labels. These can constrain unsupervised models by reducing the divergence between the expectations on target labels and on source labels or support `ambiguous learning' on the target language, respectively.



\textbf{Model transfer} instead involves training a model (e.g. a parser) on a source language and applying it on a target language \cite{zeman2008cross}, as shown in Figure \ref{fig:langtransf}b. Due to their incompatible vocabularies, models are typically delexicalized prior to transfer and take language-independent \cite{nivre-2016} or harmonized \cite{Zhang-2012} features as input. 
In order to bridge the vocabulary gap, model transfer was later augmented with multilingual Brown word clusters  \cite{Tackstrom-2012} or multilingual distributed word representations (see \S\ \ref{subsec:reprlearn}). 

\textbf{Machine translation} offers an alternative to lexicalization in absence of annotated parallel data. As shown in Figure \ref{fig:langtransf}c, a source sentence is machine translated into a target language, \citep{banea2008multilingual} or through a bilingual lexicon \citep{durrett2012syntactic}. Its annotation is then projected and used to train a target-side supervised model.
Translated documents can also be employed to generate multilingual sentence representations, which facilitate language transfer \cite{zhou2016cross}. 

Some of these methods are hampered by their resource requirements. In fact, annotation projection and translation need parallel texts to align words and train translation systems, respectively \cite{agic2015if}. Moreover, comparisons of state-of-the-art algorithms revealed that model transfer is competitive with machine translation in terms of performance \citep{conneau2018xnli}. Partly owing to these reasons, typological knowledge has been mostly harnessed in connection with model transfer, as we discuss in \S\ \ref{ssec:modelfeatures}. Moreover, typological features can guide the selection of the best source language to match to a target language for language transfer \citep[\textit{inter alia}]{agic2016multilingual}, which benefits all the above-mentioned methods (see \S\ \ref{ssec:dataselection}).

\subsection{Multilingual Joint Supervised Learning}
\label{ssec:jointlearn}

\begin{figure}[t]
\centering
\includegraphics[scale=0.65]{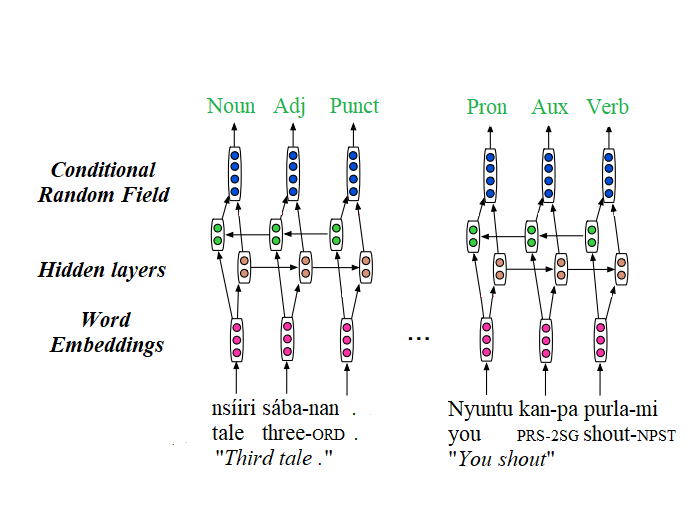}
\caption{In multilingual joint learning, representations can be private or shared across languages. Tied parameters are shown as neurons with identical color. Image adapted from \protect\namecite{fang2017model}, representing multilingual PoS tagging for Bambara (left) and Warlpiri (right).}
\label{fig:multijoint}
\end{figure}

NLP models can be learned jointly from the data in multiple languages. In addition to facilitating intrinsically multilingual applications, such as Neural Machine Translation and Information Extraction, this approach  
often surpasses language-specific monolingual models as it can leverage more (although noisier) data \cite[\textit{inter alia}]{Ammar-2016}. This is particularly true in scenarios where either a target or all languages are resource-lean \cite{Khapra-2011} or in code-switching scenarios \cite{adel2013combination}. In fact, multilingual joint learning improves over pure model transfer also in scenarios with limited amounts of labeled data in target language(s) \cite{fang2017model}.\footnote{This approach is also more cost-effective in terms of parameters \cite{pappas2017multilingual}.}



A key strategy for multilingual joint learning is \textbf{parameter sharing} \cite{johnson2016google}. More specifically, in state-of-the-art neural architectures input and hidden representations can be either private (language-specific) or shared across languages. Shared representations are the result of tying the parameters of a network component across languages, such as word embeddings \citep{guo2016exploiting}, character embeddings \citep{yang2016multi}, hidden layers \citep{duong2015neural} or the attention mechanism \citep{pappas2017multilingual}. Figure \ref{fig:multijoint} shows an example where all the components of a PoS tagger are shared between two languages (Malagasy on the left and Polish on the right).
Parameter sharing, however, does not necessarily imply parameter identity: it can be enforced by minimizing the distance between parameters \cite{duong2015low} or between latent representations of parallel sentences \cite{niehues2011wider,zhou2015subspace} in separate language-specific models. 

Another common strategy in multilingual joint modeling is providing information about the properties of the language of the current text in the form of input \textbf{language vectors} \cite{guo2016exploiting}. The intuition is that this helps tailoring the joint model toward specific languages. These vectors can be learned end-to-end in neural language modeling tasks \cite{Tsvetkov-2016,ostling2016continuous} or neural machine translation tasks \citep{johnson2016google,ha2016toward}. \namecite{Ammar-2016} instead used language vectors as a prior for language identity or typological features. 

In \S\ \ref{ssec:modelfeatures}, we discuss ways in which typological knowledge is used to balance private and shared neural network components and provide informative input language vectors. In \S\ \ref{ssec:decoding}, we argue that language vectors do not need to be limited to features extracted from typological databases, but also include automatically induced typological information \citep[see \S\ \ref{automatic-learning}]{malaviya2017learning}.

\subsection{Multilingual Representation Learning}
\label{subsec:reprlearn}

The multilingual algorithms reviewed in \autoref{ssec:langtrasf} and \autoref{ssec:jointlearn} 
are facilitated by dense real-valued vector representations of words, known as multilingual word embeddings. These can be learned from corpora and provide pivotal lexical features to several downstream NLP applications. In multilingual word embeddings, similar words (regardless of the actual language) obtain similar representations. Various methods to generate multilingual word embeddings have been developed. We follow the classification proposed by \namecite{ruder2017survey}, whereas we refer the reader to \citet{upadhyay2016cross} for an empirical comparison.

\textbf{Monolingual mapping} generates independent monolingual representations and subsequently learns a linear map between a source language and a target language based on a bilingual lexicon \cite{mikolov2013exploiting} or in an unsupervised fashion through adversarial networks \cite{conneau2017word}. 
Alternatively, both spaces can be cast into a new, lower-dimensional space through canonical correlation analysis (CCA) based on dictionaries \cite{Ammar-2016} or word alignments \cite{guo2015cross}.


\textbf{Pseudo-cross-lingual} approaches merge words with contexts of other languages and generate representations based on this mixed corpus. Substitutions are based on Wiktionary \citep{xiao2014distributed} or machine translation \cite{Gouws:2015naacl,duong2016learning}. Moreover, the mixed corpus can be produced by randomly shuffling words between aligned documents in two languages \citep{vulic2015bilingual}.

\textbf{Cross-lingual training} approaches jointly learn embeddings from parallel corpora and enforce cross-lingual constraints. This involves minimizing the distance of the hidden sentence representations of the two languages \cite{hermann2013multilingual} or decoding one from the other \cite{lauly2014learning}, possibly adding a correlation term to the loss \cite{ap2014autoencoder}. 

\textbf{Joint optimization} typically involves learning distinct monolingual embeddings, while enforcing cross-lingual constraints. These can be based on alignment-based translations \citep{Klementiev:2012coling}, 
cross-lingual word contexts \cite{luong2015bilingual}, the average representations of parallel sentences \cite{gouws2015bilbowa}, or images \cite{rotmanbridging}.

In this section, we have briefly outlined the most widely used methods in multilingual NLP. Although they offer a solution to data scarcity, cross-lingual variation remains a challenge for transferring knowledge across languages or learning from several languages simultaneously. Typological information offers promising ways to address this problem. In particular, we have noted that it can support model transfer, parameter sharing, and input biasing through language vectors. In the next two sections, we elaborate on these solutions. In particular, we review the development of typological information and the specific features which are selected for various NLP tasks (\autoref{sec:survey-development}). Afterwards, we discuss ways in which these features are integrated in NLP algorithms, for which applications they have been harnessed, and whether they truly benefit system performance (\autoref{sec:survey-applications}).

%% file: 4_development.tex
In this section we first present major publicly available typological databases and then discuss how typological information relevant to NLP models is selected, pre-processed and encoded. Finally, we highlight some limitations of database documentation with respect to coverage and feature granularity, and discuss how missing and finer-grained features can be obtained automatically.




\subsection{Hand-Crafted Documentation in Typological Databases}
\label{sec:overview-databases}

\begin{table}[th!]
\begin{center}
\def\arraystretch{1.00}
\begin{footnotesize}
\begin{tabularx}{\linewidth}{mnmX}
\toprule
{\bf Name} & {\bf Levels} & {\bf Coverage} & {\bf Feature Example} \\
\midrule
{World Atlas of Language Structures (WALS)} & {Phonology, Morphosyntax, Lexical semantics} & {2,676 languages; 192 attributes; 17\% values covered} & {\textsc{Order of Object and Verb} \newline Amele~: OV (713) \newline Gbaya Kara~: VO (705)} \\
\midrule
{Atlas of Pidgin and Creole Language Structures (APiCS)} & {Phonology, Morphosyntax} & {76 languages; 335 attributes} & {\textsc{Tense--aspect systems} \newline Ternate Chabacano~: purely aspectual (10) \newline Afrikaans~: purely temporal (1)} \\
\midrule
{URIEL Typological Compendium} & {Phonology, Morphosyntax, Lexical semantics} & {8,070 languages; 284 attributes; \textasciitilde 439,000 values} & {\textsc{Case is prefix} \newline Berber (Middle Atlas) : yes (38) \newline Hawaaian : no (993)} \\
\midrule
{Syntactic Structures of the World's Languages (SSWL)} & {Morphosyntax} & {262 languages; 148 attributes; 45\% values covered} & {\textsc{Standard negation is suffix} \newline Amharic~: yes (21) \newline Laal~: no (170)} \\
\midrule
{AUTOTYP} & {Morphosyntax} & {825 languages, \textasciitilde 1000 attributes} & {\textsc{presence of clusivity} \newline !Kung (Ju) : false \newline Ik (Kuliak) : true} \\
\midrule
{Valency Patterns Leipzig (ValPaL)} & {Predicate--argument structures} & {36 languages; 80 attributes; 1,156 values} & {\textsc{to laugh} \newline Mandinka : 1 > V \newline Sliammon : V.sbj[1] 1} \\
\midrule
{Lyon--Albuquerque Phonological Systems Database (LAPSyD)} & {Phonology} & {422 languages; \textasciitilde 70 attributes} & {\textipa{\!d} \textsc{and} \textipa{\:t} \newline Sindhi : yes (1) \newline Chuvash : no (421)} \\
\midrule
{PHOIBLE Online} & {Phonology} & {2155 languages; 2,160 attributes} & {m \newline Vietnamese : yes (2053) \newline Pirahã : no (102)} \\
\midrule
{StressTyp2} & {Phonology} & {699 languages; 927 attributes} & {\textsc{stress on first syllable} \newline Koromfé : yes (183) \newline Cubeo : no (516)} \\
\midrule
{World Loanword Database (WOLD)} & {Lexical semantics} & {41 languages; 24 attributes; \textasciitilde 2000 values} & {\textsc{horse} \newline Quechua~: \textit{kaballu} borrowed (24) \newline Sakha~: \textit{s\textipa{1}lg\textipa{1}} no evidence (18)} \\
\midrule
{Intercontinental Dictionary Series (IDS)} & {Lexical semantics} & {329 languages; 1310 attributes} & {\textsc{world} \newline Russian : \textit{mir} \newline Tocharian A : \textit{\=arki\'so\d si}} \\
\midrule
{Automated Similarity Judgment Program (ASJP)} & {Lexical Semantics} & {7,221 languages; 40 attributes} & {I \newline Ainu Maoka : \textit{co7okay} \newline Japanese : \textit{watashi}} \\
\bottomrule
\end{tabularx}
\end{footnotesize}
\end{center}
\vspace{-0.0em}
\caption{An overview of major publicly accessible databases of typological information. The databases are ordered by description level (and secondly by date of creation), along with their coverage. The table also provides feature examples: for each feature (in small capitals) we present two example languages with distinct feature values, and the total number of languages with each value in parenthesis (where applicable).}

\vspace{-0.1em}
\label{tab:databases}
\end{table}

\begin{figure}[ht]
\includegraphics[width=\textwidth]{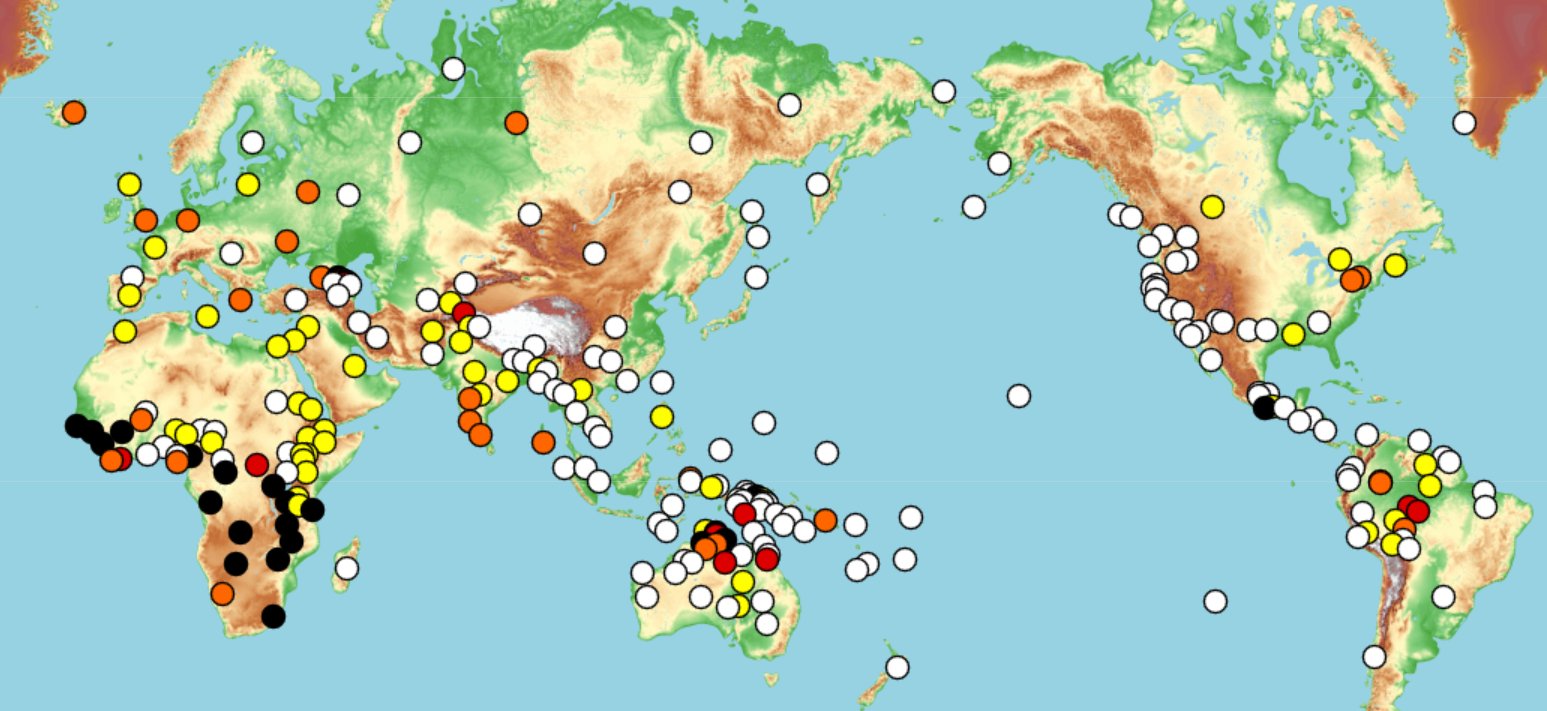}
\caption{Number of grammatical genders for nouns in the world's languages according to WALS \protect{\citep{wals-2013}}: none (white), two (yellow), three (orange), four (red), five or more (black).}
\label{fig:gendermap}
\end{figure}

Typological databases are created manually by linguists. They contain taxonomies of typological features, their possible values, as well as the documentation of feature values for the world's languages. 
 Major typological databases, listed in Table~\ref{tab:databases}, typically organize linguistic information in terms of universal features and language-specific values. For example, Figure \ref{fig:gendermap} presents language-specific values for the feature \emph{number of grammatical genders for nouns} on a world map. Note that each language is color-coded according to its value. Further examples for each database can be found in the rightmost column of Table~\ref{tab:databases}.

Some databases store information pertaining to multiple levels of linguistic description. These include the World Atlas of Language Structures (WALS) \cite{wals-2013} and the Atlas of Pidgin and Creole Language Structures (APiCS) \cite{apics}. Among all presently available databases, WALS has been the most widely used in NLP. In this resource, which has 142 typological features in total, 1--19 deal with phonology, 20--29 with morphology, 30--57 with nominal categories, 58--64 with nominal syntax, 65--80 with verbal categories, 81--97 and 143--144 with word order, 98--121 with simple clauses, 122--128 with complex sentences, 129--138 with the lexicon, and 139--142 with other properties. 


Other databases only cover features at a specific level of linguistic description. For example, both Syntactic Structures of the World's Languages (SSWL) \cite{sswl} and AUTOTYP \citep{autotyp} focus on syntax. SSWL features are manually crafted, whereas AUTOTYP features are derived automatically from primary linguistic data using scripts. The Valency Patterns Leipzig (ValPaL) \cite{valpal} provides verbs as attributes and predicate--argument structures as their values (including both valency and morphosyntactic constraints). For example, in both Mandinka and Sliammon, the verb \textit{to laugh} has a valency of 1; in other words, it requires only one mandatory argument, the subject. In Mandinka the subject precedes the verb, but there is no agreement requirement; in Sliammon, on the other hand, the word order does not matter, but the verb is required to morphologically agree with the subject. 

For phonology, the Phonetics Information Base and Lexicon (PHOIBLE) \cite{phoible} collates information on segments (binary phonetic features). In the Lyon--Albuquerque Phonological Systems Database (LAPSyD) \cite{Maddieson-2013}, attributes are articulatory traits, syllabic structures or tonal systems. Finally, StressTyp2 \cite{stresstyp2} deals with stress and accent patterns. For instance, in Koromfé each word's first syllable has to be stressed, but not in Cubeo.

Other databases document various aspects of semantics. The World Loanword Database (WOLD) \cite{wold} documents loanwords by identifying the donor languages and the source words. The Automated Similarity Judgment Program (ASJP) \cite{asjp} and the Intercontinental Dictionary Series (IDS) \cite{ids} indicate how a meaning is lexicalized across languages: e.g. the concept of \textsc{world} is expressed as \textit{mir} in Russian, and as \textit{\=arki\'so\d si} in Tocharian A.

Although typological databases store abundant information on many languages, they suffer from shortcomings that limit their usefulness. Perhaps the most significant shortcoming of such resources is their limited coverage. In fact, feature values are missing for most languages in most databases. Other shortcomings are related to feature granularity. In particular, most databases fail to account for feature value variation within each language: they report only majority value rather than the full range of possible values and their corresponding frequencies. For example, the dominant Adjective--Noun word order in Italian is Adjective before Noun; however, the opposite order is also attested. The latter information is often missing from typological databases.

Further challenges are posed by restricted feature applicability and feature hierarchies. Firstly, some features apply, by definition, only to subsets of languages that share another feature value. For instance, WALS feature 113A documents ``Symmetric and Asymmetric Standard Negation", whereas WALS feature 114A ``Subtypes of Asymmetric Standard Negation''. Although a special NA value is assigned for symmetric-negation languages in the latter, there are cases where languages without the prerequisite feature are simply omitted from the sample. 
Secondly, features can be partially redundant, and subsume other features. For instance, WALS feature 81A ``Order of Subject, Object and Verb'' encodes the same information as WALS feature 82A ``Order of Subject and Verb'' and 83A ``Order of Object and Verb'', with the addition of the order of Subject and Object.

\subsection{Feature Selection from Typological Databases}
\label{sec:manual-extraction}

\begin{figure}[t]
\centering
\includegraphics[width=\textwidth]{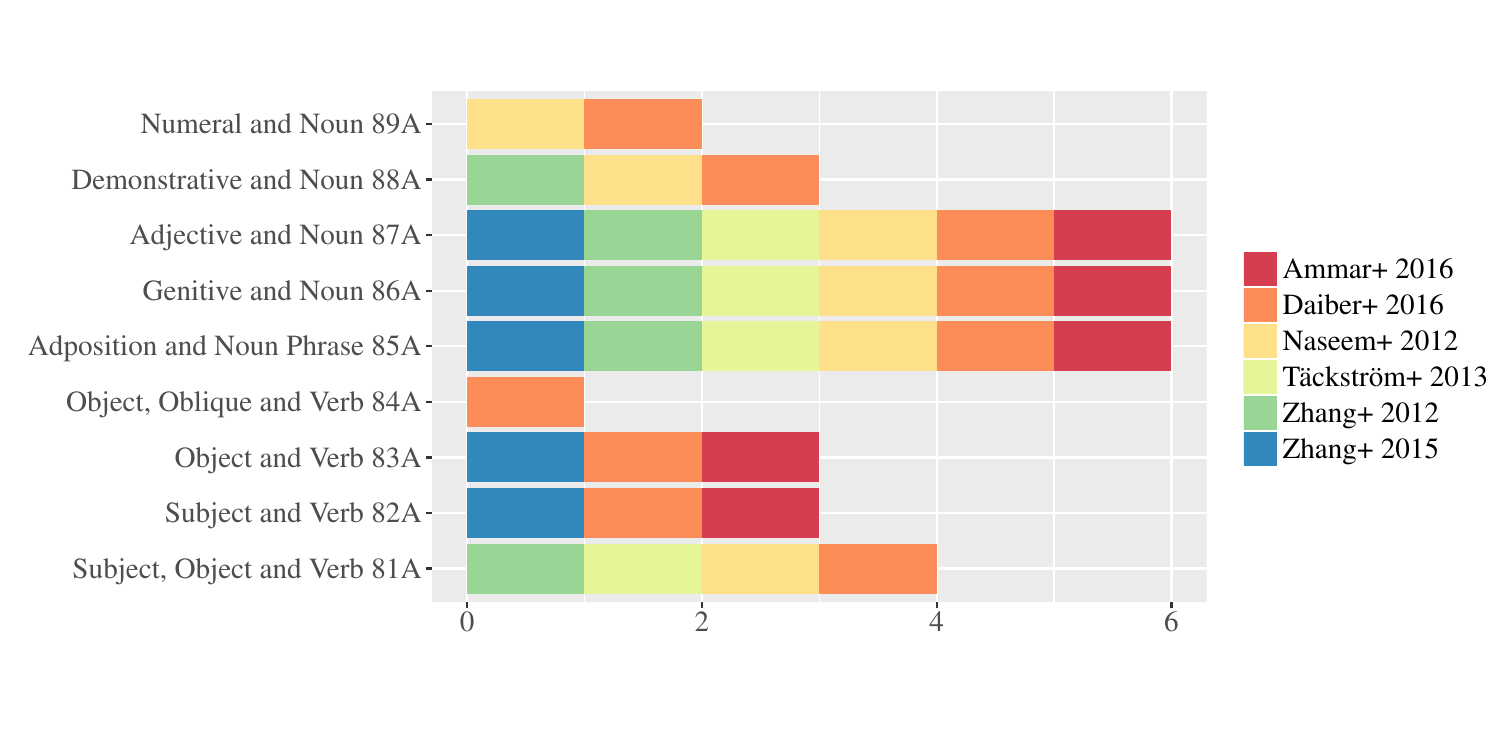}
\caption{Feature sets employed in a sample of typologically informed experiments for dependency parsing. The numbers refer to WALS ordering \protect{\citep{wals-2013}}.}
\label{fig:walsfeats}
\end{figure}

The databases presented above can serve as a rich source of typological information for NLP. In this section, we survey the feature sets that have been extracted from these databases in typologically informed NLP studies. In \autoref{ssec:comptask}, we review in which ways and to which degree of success these features have been integrated in machine learning algorithms.

Most NLP work incorporated a subset of word order features from WALS \citep{wals-2013}, mostly for the task of syntactic dependency parsing, where word order provides crucial guidance \cite{Naseem-2012}. The feature subsets used in different studies are shown in Figure~\ref{fig:walsfeats}. As depicted in the figure, these studies employed quite similar word order features. The feature set first established by \namecite{Naseem-2012} served as inspiration for subsequent works. These works, however, often discarded features with the same value for all of the languages in their sample, as these were not discriminative. 

Another group of studies used more comprehensive feature sets. The feature set of \citet{daiber2016universal} included not only WALS word order features but also nominal categories (e.g.\ `Conjunctions and Universal Quantifiers') and nominal syntax (e.g.\ `Possessive Classification'). \namecite{Berzak-2015} considered all features from WALS associated with morphosyntax and pruned out the redundant ones, resulting in a total of 119 features. \namecite{sogaard2012empirical} utilized all the features in WALS with the exception of phonological features. \namecite{Tsvetkov-2016} selected 190 binarized phonological features from URIEL \citep{Littel-2016}. These features encoded the presence of single segments, classes of segments, minimal contrasts in a language inventory, and the number of segments in a class. For instance, they record whether a language allows two sounds to differ only in voicing, such as /t/ and /d/.

Finally, a small number of experiments adopted the entire feature inventory of typological databases, without any sort of pre-selection. In particular \namecite{agic2017cross} and \namecite{Ammar-2016} extracted all the features in WALS, while \citep{Deri-2016} all the features in URIEL. \citet{schone2001language} did not resort to basic typological features, but rather to ``several hundred [implicational universals] applicable to syntax'' drawn from the Universal Archive \citep{Plank-1996}. 

Typological attributes that are extracted from typological databases are typically represented as feature vectors in which each dimension encodes a feature value. This feature representation is often binarized \cite{Georgi-2010}: for each possible value \textit{v} of each database attribute \textit{a}, a new feature is created with value 1 if it corresponds to the actual value for a specific language and 0 otherwise. Note that this increases the number of features by a factor of $\frac{1}{||a||} \sum_{i=1}^{||a||} ||v_{a_i}||$. Although binarization helps harmonizing different features and different databases, it obscures the different types of typological variables.

\begin{figure}[t!]
    \centering
    \begin{subfigure}[t]{0.49\textwidth}
        \includegraphics[width=\linewidth]{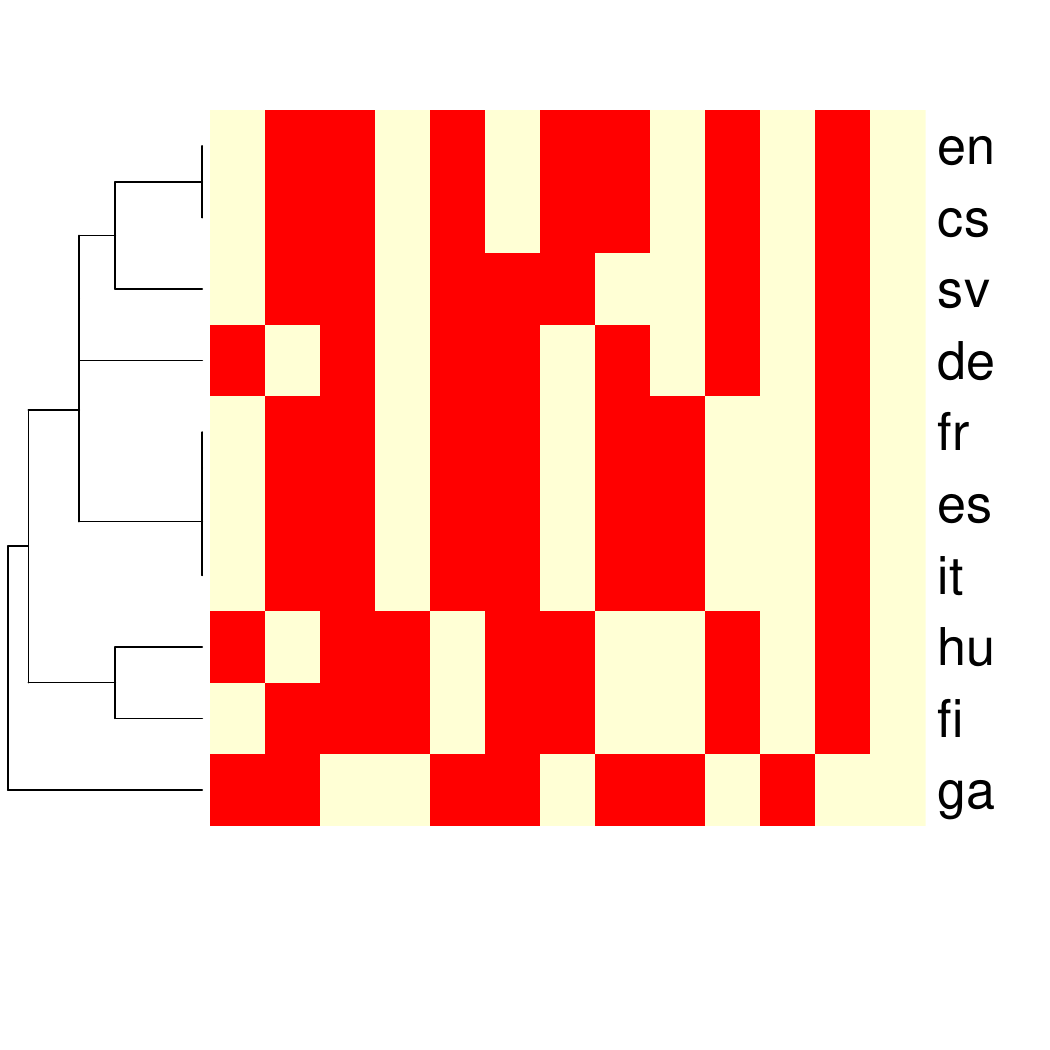}
  \caption{Word-order features}\label{fig:heatmap_wo}
    \end{subfigure}%
    ~ 
    \begin{subfigure}[t]{0.49\textwidth}
        \includegraphics[width=\linewidth]{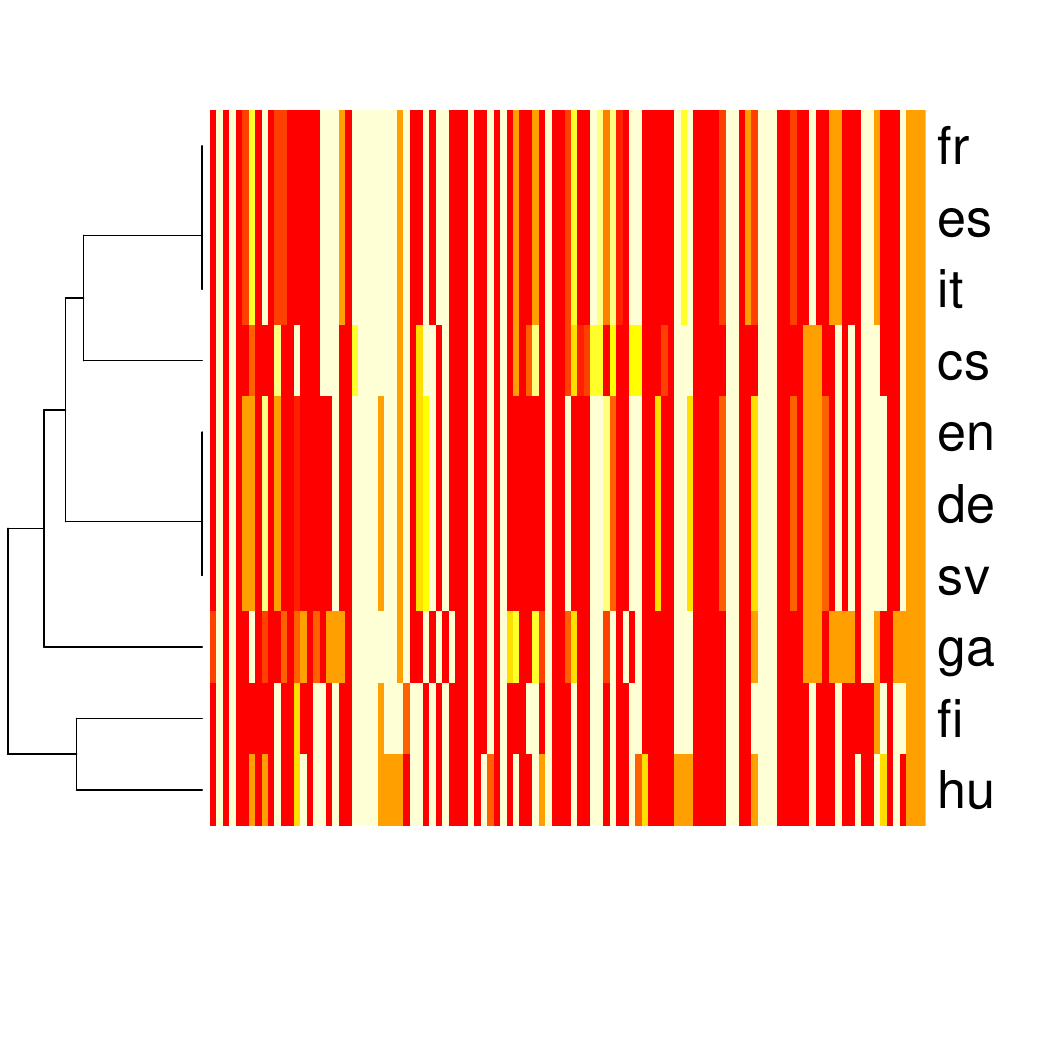}
  \caption{All WALS features with average genus values}\label{fig:heatmap_all}
    \end{subfigure}
    \caption{Heat maps of encodings for different subsets of typological WALS features taken from \protect{\citet{Ammar-2016}}: rows stand for languages, dimensions for attributes, and color intensities for feature values. Encodings are clustered hierarchically by similarity. The meaning of language codes is: \textsc{de} German, \textsc{cs} Czech, \textsc{en} English, \textsc{es} Spanish, \textsc{fr} French, \textsc{fi} Finnish, \textsc{ga} Irish Gaelic, \textsc{hu} Hungarian, \textsc{it} Italian, \textsc{sv} Swedish.}
    \label{fig:lanreps}
\end{figure}

\begin{table}[p]
\begin{center}
\def\arraystretch{1.00}
\begin{footnotesize}
\begin{tabularx}{\linewidth}{smmnsm}
\toprule
 & {\bf Author} & {\bf Details} & {\bf Requirements} & {\bf Langs} & {\bf Features} \\
\midrule

\parbox[t]{2mm}{\multirow{5}{*}{\rotatebox[origin=c]{90}{Morphosyntactic Annotation}}} & \citet{liu2010dependency} & Treebank count & Treebank & 20 & word order\\
\cline{2-6}
& \citet{Lewis-2008} & IGT projection & IGT, source chunker & 97 & word and morpheme order, determiners\\
& \citet{Bender-2013} &  IGT projection & IGT, source chunker & 31 & word order and case alignment\\
\cline{2-6}
& \citet{Ostling-2015} & Treebank projection & Parallel text, source tagger and parser & 986 & word order\\
& \citet{Zhang-2016} & PoS projection & source tagger, seed dictionary & 6 & word order\\

\midrule
\parbox[t]{2mm}{\multirow{7}{*}{\rotatebox[origin=c]{90}{Unsupervised Propagation}}} & \citet{teh2007bayesian} & Hierarchical typological cluster & WALS & 2150 & whole \\
& \namecite{Georgi-2010} & Majority value from k-means typological cluster & WALS & whole & whole \\
& \citet{Coke-2016} & Majority value from genus & Genealogy and WALS & 325 & word order and passive\\
\cline{2-6}
& \citet{Littel-2016} & Family, area, and typology-based Nearest Neighbors & Genealogy and WALS & whole & whole\\
& \citet{Berzak-2014} & English as a Second Language-based Nearest Neighbors & ESL texts & 14 & whole \\
\cline{2-6}
& \citet{malaviya2017learning} & Task-based language vector & NMT dataset & 1017 & whole\\
& \citet{bjerva2018phonology} & Task-based language vector & PoS tag dataset & 27-824 & phonology, morphology, syntax\\

\midrule
\parbox[t]{2mm}{\multirow{6}{*}{\rotatebox[origin=c]{90}{Supervised Learning}}} & \citet{Takamura-2016} & Logistic regression & WALS & whole & whole\\
& \citet{murawaki2017diachrony} & Bayesian + feature and language interactions & Genealogy and WALS & 2607 & whole\\
& \citet{wang-eisner-2017} & Feed-forward Neural Network & WALS, tagger, synthetic treebanks & 37 & word order\\
& \citet{cotterell2017probabilistic} & Determinant Point Process with neural features & WALS & 200 & vowel inventory\\
\cline{2-6}
& \citet{Daume-2007} & Implication universals & Genealogy and WALS & whole & whole \\
& \citet{lu2013exploring} & Automatic discovery & Genealogy and WALS & 1646 & word order \\

\midrule
\parbox[t]{2mm}{\multirow{3}{*}{\rotatebox[origin=c]{90}{\shortstack[l]{Cross-lingual \\ distribution}}}} & 
\citet{Walchli:2012ling} & Sentence edit distance  & Multi-parallel texts, pivot & 100 & motion verbs \\
& \citet{asgari2017past} & Pivot alignment & Multi-parallel texts, pivot & 1163  & tense markers \\
\cline{2-6}
& \citet{roy2014automatic} & Correlations in counts and entropy & None & 23 & adposition word order \\

\bottomrule
\end{tabularx}
\end{footnotesize}
\end{center}
\vspace{-0.0em}
\caption{An overview of the strategies for prediction of typological features.}
\vspace{-0.1em}
\label{tab:strategies}
\end{table}

To what extent do the limitations of typological databases mentioned in \autoref{sec:overview-databases} affect the feature sets surveyed in this section? The coverage is generally broad for the languages used in these experiments, as they tend to be well documented. For instance, on average 79.8\% of the feature values are populated for the 14 languages appearing in \citet{Berzak-2015}, as opposed to 17 percent for all the languages in WALS.

It is hard to assess at a large scale how informative is a set of typological features. However, these can be meaningfully compared with genealogical information. Ideally, these two properties should not be completely equivalent (otherwise they would be redundant),\footnote{This does not apply to isolates, however: by definition, no genealogical information is available for these languages. Hence, typology is the only source of information about their properties.} but at the same time they should partly overlap (language cognates inherit typological properties from the same ancestors). In Figure \ref{fig:lanreps}, we show two feature sets appearing in \citet{Ammar-2016}, each depicted as a heatmap. Each row represents a language in the data, each cell is colored according to the feature value, ranging from 0 to 1. In particular, the feature set of Figure \ref{fig:heatmap_wo} is the subset of word order features listed in Figure~\ref{fig:walsfeats}; and Figure \ref{fig:heatmap_all} is a large set of WALS features where values are averaged by language genus to fill in missing values.

In order to compare the similarities of the typological feature vectors among languages, we clustered languages hierarchically based on such vectors.\footnote{Clustering was performed through the complete linkage method.} Intuitively, the more this hierarchy resembles their actual family tree, the more redundant typological information is. This is the case for \ref{fig:heatmap_all}, where the lowest-lever clusters correspond exactly to a genus or family (top-down: Romance, Slavic, Germanic, Celtic, Uralic). Still, the language vectors belonging to the same cluster display some micro-variations in individual features. On the other hand, \ref{fig:heatmap_wo} shows clusters differing from language genealogy: for instance, English and Czech are merged although they belong to different genera (Germanic and Slavic). However, this feature set fails to account for fine-grained differences among related languages: for instance, French, Spanish, and Italian receive the same encoding.\footnote{Notwithstanding they have different \textit{preferences} over word orders \citep{liu2010dependency}.}


To sum up, this section's survey on typological feature sets reveals that most experiments have taken into account a small number of databases and features therein. However, several studies did utilize a larger set of coherent features or full databases. Although well-documented languages do not suffer much from coverage issues, we showed how difficult it is to select typological features that are non-redundant with genealogy, fully discriminative, and informative. The next section addresses these problems proposing automatic prediction as a solution.

\subsection{Automatic Prediction of Typological Features}
\label{automatic-learning}

The partial coverage and coarse granularity of existing typological resources sparked a line of research on automatic acquisition of typological information. Missing feature values can be predicted based on: i) heuristics from morphosyntactic annotation that pre-exists, such as treebanks, or is transferred from aligned texts (\S\ \ref{sssec:morsynann}); ii) unsupervised propagation from other values in a database based on clustering or language similarity metrics (\S\ \ref{sssec:propdat}); iii) supervised learning with Bayesian models or artificial neural networks (\S\ \ref{sssec:suppre}); or iv) heuristics based on co-occurrence metrics, typically applied to multi-parallel texts (\S\ \ref{sssec:multali}). These strategies are summarized in Table \ref{tab:strategies}. 

With the exception of \citet{Naseem-2012}, who treated typological information as a latent variable, automatically acquired typological features have not been integrated into algorithms for NLP applications to date. However, they have several advantages over manually crafted features. Unsupervised propagation and supervised learning fill in missing values in databases, thereby extending their coverage. Moreover, heuristics based on morphosyntactic annotation and co-occurrence metrics extract additional information that is not recorded in typological databases. Further, they can account for the distribution of feature values within single languages, rather than just the majority value. Finally, they do not make use of discrete cross-lingual categories to compare languages; rather, language properties are reflected in continuous representations, which is in line with their  gradient nature (see \autoref{sec:overview})

\subsubsection{Heuristics based on morphosyntactic annotation}
\label{sssec:morsynann}
\begin{figure}[th!]
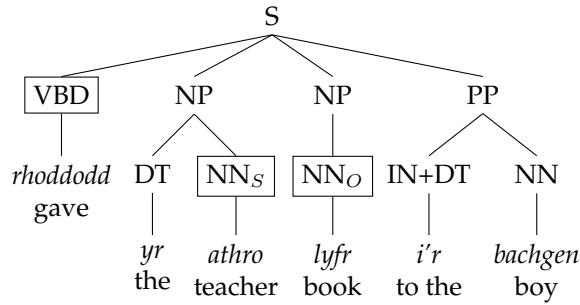

\centering
\tikzset{every tree node/.style={align=center,anchor=north}}
\Tree [.S [.\node[draw]{VBD}; {\textit{rhoddodd} \\ gave} ] [.NP [.DT {\textit{yr} \\ the} ] [.\node[draw]{NN$_S$}; {\textit{athro} \\ teacher} ] ] [.NP [.\node[draw]{NN$_O$}; {\textit{lyfr} \\ book} ] ] [.PP [.IN+DT {\textit{i'r} \\ to the} ] [.NN {\textit{bachgen} \\ boy} ] ]]
\caption{Constituency tree of a Welsh sentence.}
\label{fig:welshtree}
\end{figure}

Morphosyntactic feature values can be extracted via heuristics from morphologically and syntactically annotated texts. For example, word order features can be calculated by counting the average direction of dependency relations or constituency hierarchies \citep{liu2010dependency}. Consider the tree of a sentence in Welsh from \namecite{Bender-2013} in Figure \ref{fig:welshtree}. The relative order of verb--subject, and verb--object can be deduced from the position of the relevant nodes \textit{VBD}, \textit{NN$_S$} and \textit{NN$_O$} (highlighted).

\noindent
Morphosyntactic annotation is often unavailable for resource-lean languages. In such cases, it can be projected from a source directly to several target languages through language transfer. For instance,
\namecite{Ostling-2015} project source morphosyntactic annotation directly to several languages through a multilingual word alignment. After the alignment and projection, word order features are calculated by the average direction of dependency relations. Similarly, \namecite{Zhang-2016} transfer POS annotation with a model transfer technique relying on multilingual embeddings, created through monolingual mapping (see \S\ \ref{subsec:reprlearn}). After the projection, they predict feature values with a multiclass Support Vector Machine (SVM) using POS tag n-gram features.

Finally, typological information can be extracted from Interlinear Glossed Texts (IGT). Such collections of example sentences are collated by linguists and contain grammatical glosses with morphological information. These can guide alignment between the example sentence and its English translation. \namecite{Lewis-2008} and \namecite{Bender-2013} project chunking information from English and train Context Free Grammars on target languages. After collapsing identical rules, they arrange them by frequency and infer word order features.

\subsubsection{Unsupervised propagation}
\label{sssec:propdat}

Another line of research seeks to increase the coverage of typological databases borrowing missing values from the known values in other languages. One approach is clustering languages according to some criterion and propagating the majority value within each cluster. 
Hierarchical clusters can be created either according to typological features (e.g. \namecite{teh2007bayesian}) or 
based on language genus \citep{Coke-2016}. Through extensive evaluation, \namecite{Georgi-2010} demonstrate that typology based clustering outperforms genealogical clustering for unsupervised propagation of typological features. Among the clustering techniques examined, k-means appears to be the most reliable as compared to k-medoids, the Unweighted Pair Group Method with Arithmetic mean (UPGMA), repeated bisection, and hierarchical methods with partitional clusters.


\begin{figure}[t]
    \centering
    \begin{subfigure}[t]{0.27\textwidth}
        \centering
        \includegraphics[width=\textwidth]{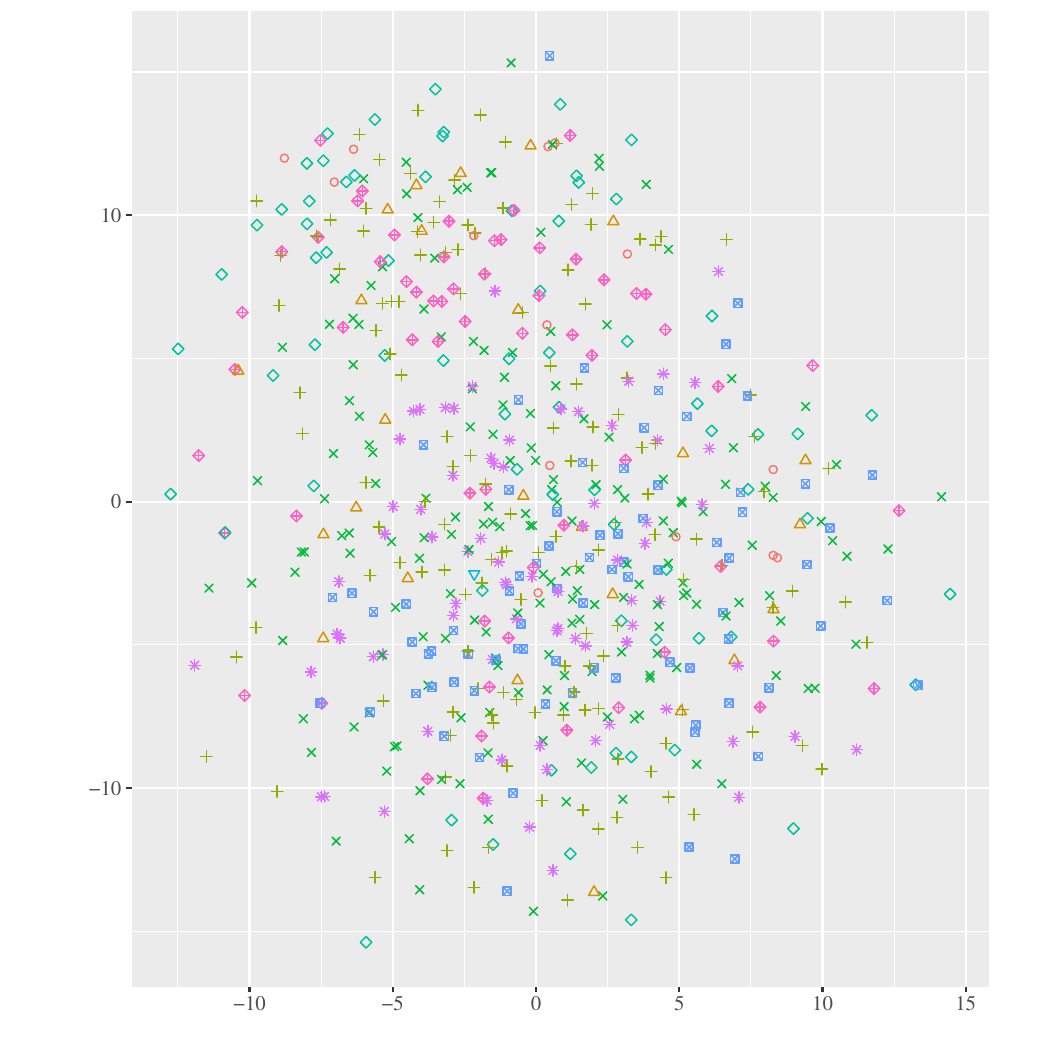}
        \caption{Input vectors}
        \label{tsne1}
    \end{subfigure}%
    ~ 
    \begin{subfigure}[t]{0.27\textwidth}
        \centering
        \includegraphics[width=\textwidth]{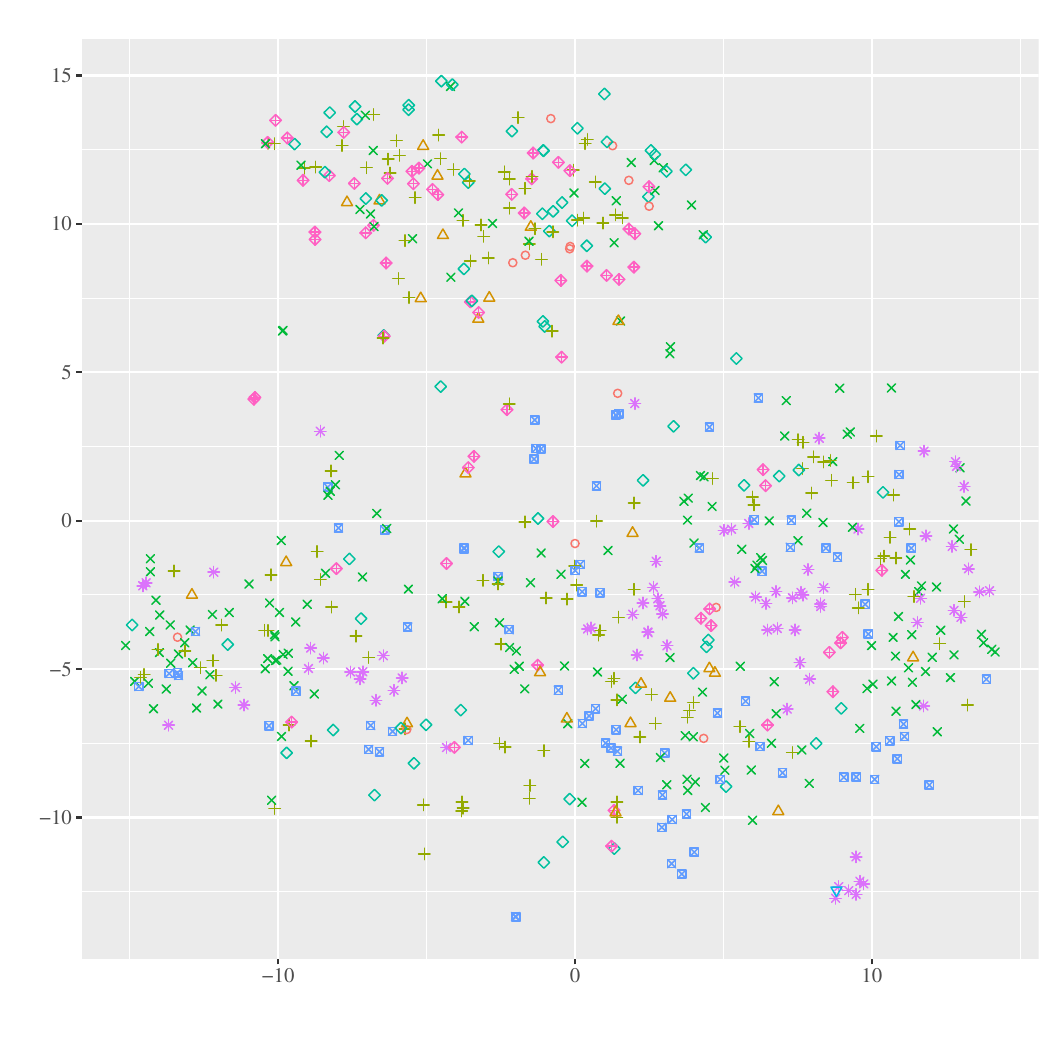}
        \caption{Cell states}
        \label{tsne2}
    \end{subfigure}%
    ~ 
    \begin{subfigure}[t]{0.44\textwidth}
        \centering
        \includegraphics[clip,trim={0 3.5cm 0 3.5cm},width=\textwidth]{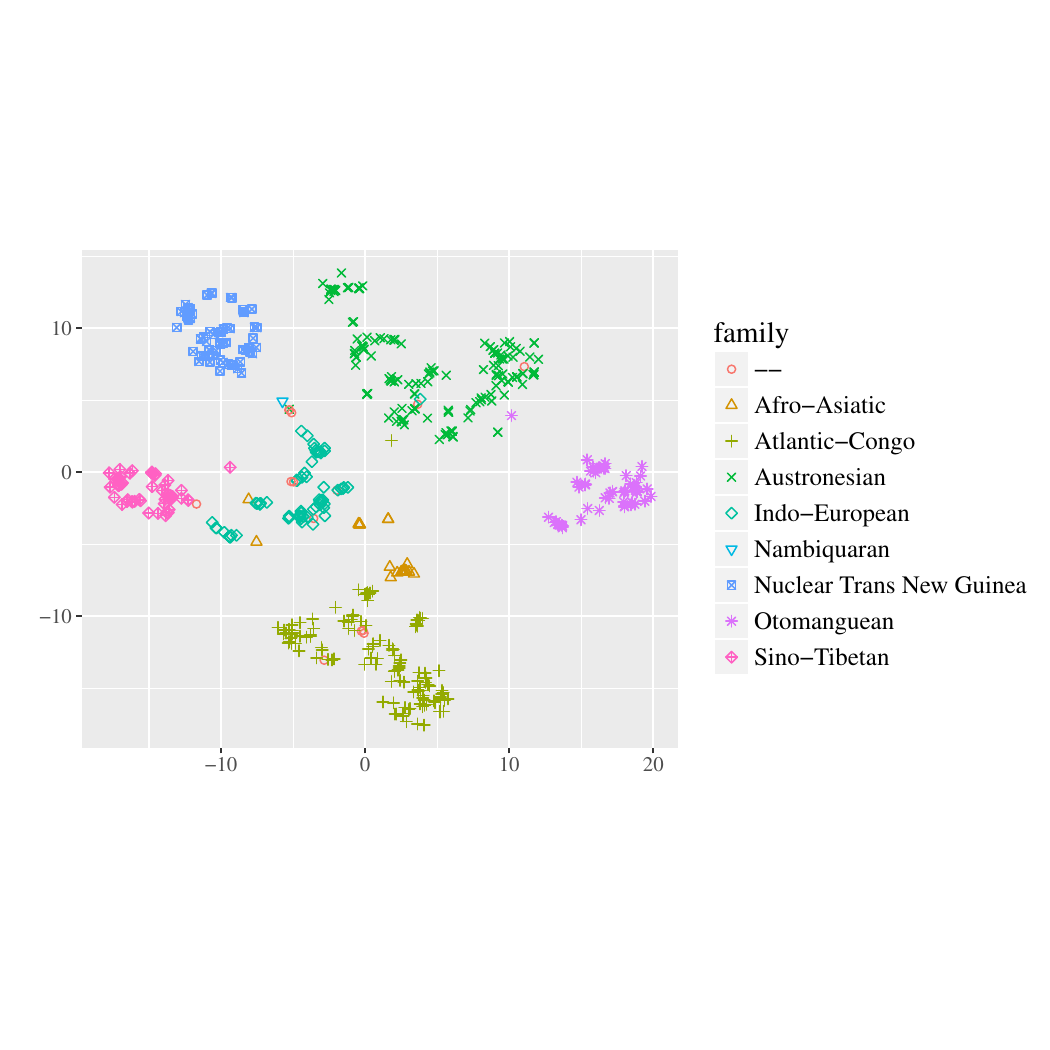}
        \caption{Predicted WALS}
        \label{tsne3}
    \end{subfigure}%
    ~
    \caption{Language representations dimensionality-reduced with t-SNE.}
    \label{tsne}
\end{figure}

Language similarity measures can also rely on a distributed representation of each language. These language vectors are trained end-to-end as part of neural models for downstream tasks such as many-to-one Neural Machine Translation. 
In particular, language vectors can be obtained from artificial trainable tokens concatenated to every input sentence, similarly to \citet{johnson2016google}, or from the aggregated values of the hidden state of a neural encoder. Using these language representations, typological feature values are propagated using K Nearest Neighbors \citep{bjerva2018phonology} or predicted with logistic regression \citep{malaviya2017learning}. 



Language vectors can be conceived as data-driven, continuous typological representations of a language, and as such provide an alternative to manually crafted typological representations.
Similarly to the analysis carried out in \autoref{sec:manual-extraction}, we can investigate how much language vectors align with genealogical information. Figure \ref{tsne} compares continuous representations based on artificial tokens (Figure \ref{tsne1}) and encoder hidden states (Figure \ref{tsne2}) with vectors of discrete WALS features from URIEL (Figure \ref{tsne3}). All the representations are reduced to 2 dimensions with \textit{t}-SNE, and color-coded based on their language family.

As the plots demonstrate, the information encoded in WALS vectors is akin to genealogical information, partly because of biases introduced by family-based propagation of missing values \citep[see \S\ \ref{sssec:propdat}]{Littel-2016}. On the other hand, artificial tokens and encoder hidden states cannot be reduced to genealogical clusters. Yet, their ability to predict missing values is not inferior to WALS features (as detailed in \S\ \ref{ssec:typrecom}). This implies that discrete and continuous representations appear to capture different aspects of the cross-lingual variation, while being both informative. For this reason, they are possibly complementary and could be combined in the future.

\subsubsection{Supervised learning}
\label{sssec:suppre}

As an alternative to unsupervised propagation, one can learn an explicit model for predicting feature values through supervised classification. For instance, \citet{Takamura-2016} use logistic regression with WALS features and evaluate this model in a cross-validation setting where one language is held out in each fold. \citet{wang-eisner-2017} provide supervision to a feed-forward neural network with windows of PoS tags from natural and synthetic corpora. 

Supervised learning of typology can also be guided by non typological information (see \autoref{sec:overview}). Within the Bayesian framework, \citet{murawaki2017diachrony} exploits not only typological but also genealogical and areal dependencies among languages to represent each language as a binary latent parameter vector through a series of autologistic models. 
\citet{cotterell2017probabilistic,cotterell2018deep} develop a point-process generative model of vowel inventories (represented as either IPA symbols or acoustic formants) based on some universal cognitive principles: dispersion (phonemes are as spread out as possible in the acoustic space) and focalization (some positions in the acoustic space are preferred due to the similarity of the main formants). 

An alternative approach to supervised prediction of typology is based on learning implicational universals of the kind pioneered by \namecite{Greenberg-1963} with probabilistic models from existing typological databases. Using such universals, features can be deduced by \textit{modus ponens}. For instance, once it has been established that the presence of `High consonant/vowel ratio' and `No front-rounded vowels' implies `No tones', the latter feature value can be deduced from the premises if those are known. \namecite{Daume-2007} proposes a Bayesian model for learning typological universals that predicts features based on the intuition that their likelihood does not equal their prior probability, but rather is constrained by other features. \citet{lu2013exploring} cast this problem as knowledge discovery, where language features are encoded in a Directed Acyclic Graph. The strength of implication universals is represented as weights associated with the edges of this graph.



\begin{figure}[b]
\centering
\includegraphics[scale=0.4]{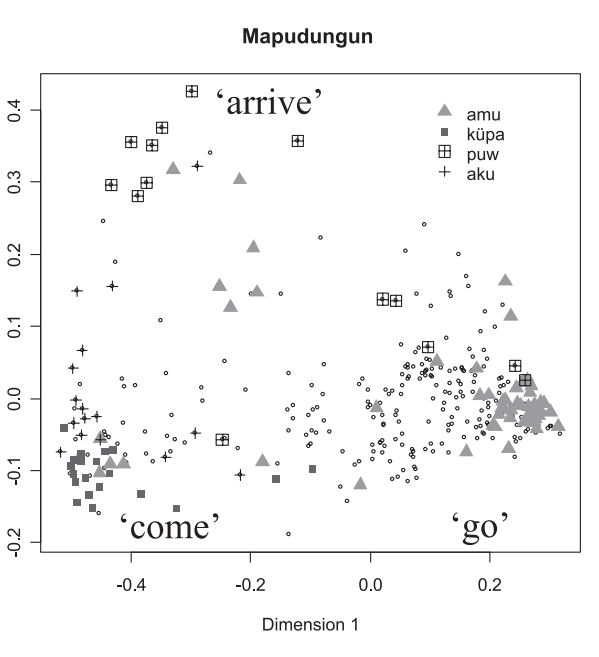}
\caption{\protect{\citet{Walchli:2012ling}}'s cross-lingual sentence visualization for Mapungundun. In the top-right corner is the legend of the motion verbs taken into consideration. Each data point is an instance of a verb in a sentence, positioned according to its contextualized sense. English glosses are the authors' interpretations of the main clusters.}
\label{fig:cysouw}
\end{figure}

\subsubsection{Heuristics based on Cross-Lingual Distributional Information}
\label{sssec:multali}
Typological features can also emerge in a data-driven fashion, based on distributional information from multi-parallel texts.
\citet{Walchli:2012ling} create a matrix where each row is a parallel sentence, each column is a language, and cell values are lemmas of motion verbs occurring in those sentences. This matrix can be transformed to a (Hamming) distance matrix between sentence pairs, and reduced to lower dimensionality via Multi-Dimensional Scaling (MDS). This provides a continuous map of lexical semantics that is language-specific, but motivated by categories that emerge across languages. For instance, Figure \ref{fig:cysouw} shows the first two dimensions of the MDS similarity matrix in Mapudungun, where the first dimension can be interpreted as reflecting the direction of motion.


\citet{asgari2017past} devised a procedure to obtain markers of grammatical features across languages. Initially, they manually select a language containing an unambiguous and overt marker for a specific typological feature (called head pivot) based on linguistic expertise. For instance, \textit{ti} in Seychellois Creole (French Creole) is a head pivot for past tense marking. Then, this marker is connected to equivalent markers in other languages through alignment-based $\chi^2$ test in a multi-parallel corpus and n-gram counts. 

Finally, typological features can be derived from raw texts in a completely unsupervised fashion, without multi-parallel texts. \namecite{roy2014automatic} use heuristics to predict the order of adpositions and nouns. Adpositions are identified as the most frequent words. Afterwards, the position of the noun is established based on whether selectional restrictions appear on the right context or the left context of the adposition, according to count-based and entropy-based metrics.

\subsubsection{Comparison of the strategies}
\label{ssec:typrecom}

\begin{figure}[t]
\vspace{-2cm}
    \centering
    \includegraphics[width=0.9\textwidth]{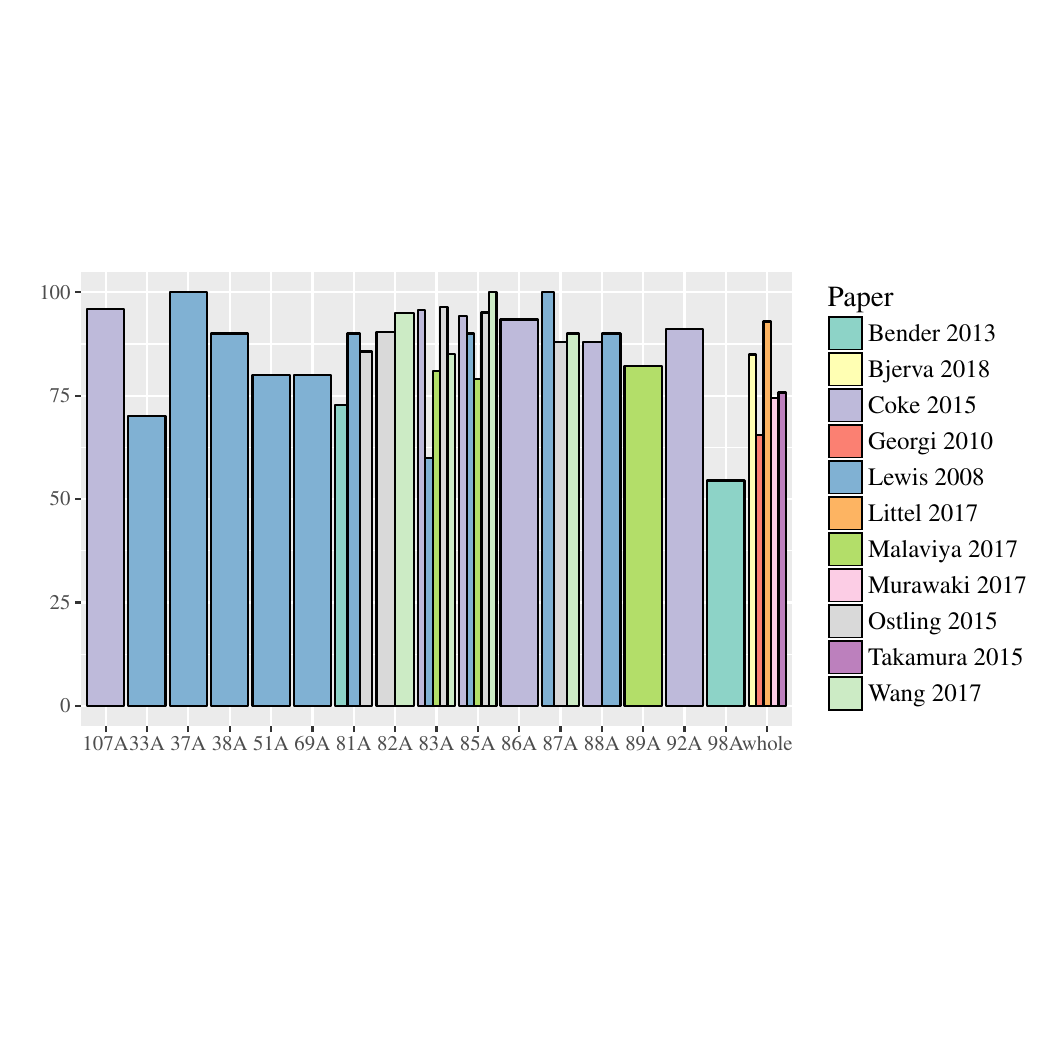}
    \vspace{-2cm}

    \caption{Accuracy of different approaches (see legend on the right) in predicting missing values of WALS typological features (specified on the vertical axis).}
    \label{fig:predres}

\end{figure}

Establishing which of the above-mentioned strategies is optimal in terms of prediction accuracy is not straightforward. In Figure \ref{fig:predres}, we collect the scores reported by several of the surveyed papers, provided that they concern specific features or the whole WALS dataset (as opposed to subsets) and are numerical (as opposed to graphical plots). However, these results are not strictly comparable, since language samples and/or the split of data partitions may differ. The lack of standardization in this respect allows us to draw conclusions only about the difficulty of predicting each feature relative to a specific strategy: for instance, the correct value of passive voice is harder to predict than word order, as claimed by \citet{Bender-2013} and witnessed by Figure \ref{fig:predres}.

However, some papers carry out comparisons of the different strategies within the same experimental setting. According to \namecite{Coke-2016}, propagation from the genus majority value outperforms logistic regression among word-order typological features. On the other hand, \namecite{Georgi-2010} argue that typology-based clusters are to be preferred in general. This apparent contradiction stems from the nature of the target features: genealogy excels in word order features due to their diachronic stability. As they tend to be preserved over time, they are often shared by all members of a family. In turn, majority value propagation is surpassed by supervised classification when evaluated on the entire WALS feature set \citep{Takamura-2016}.

In general, there appears to be no `one-size-fits-all' algorithm. For instance, \citet{Coke-2016} outperform \citet{wang-eisner-2017} for object--verb order (83A) but are inferior to it for adposition--noun (85A).
In fact, each strategy is suited for different features, and requires different resources. Based on Figure \ref{fig:predres}, the extraction of information from morphosyntactic annotation is well suited for word order features, whereas distributional heuristics from multi-parallel texts are more informative about lexicalization patterns. On the other hand, unsupervised propagation and supervised learning are general-purpose strategies. Moreover, the first two presuppose some annotated and/or parallel texts, whereas the second two need a pre-existing database documentation. Strategies may be preferred according to which resources are available for a specific language.

Many strategies have a common weakness, however, as they postulate incorrectly that language samples are independent and identically distributed \citep{lu2013exploring,cotterell2017probabilistic}. This is not the case due to the interactions of family, area, and implicational universals. The solutions adopted to mitigate this weakness vary: \citet{wang-eisner-2017} balance the data distribution with synthetic examples, whereas \citet{Takamura-2016} model family and area interactions explicitly. However, according to \citet{murawaki2017diachrony}, these interactions have different degrees of impact on typological features. In particular, inter-feature dependencies are more influential than inter-language dependencies, and horizontal diffusibility (borrowing from neighbours) is more prominent than vertical stability (inheriting from ancestors).

Finally, a potential direction for future investigation emerges from this section's survey. In addition to missing value completion, automatic prediction often also accounts for the variation internal to each language. However, some strategies go even further, and ``open the way for a typology where generalizations can be made without there being any need to reduce the attested diversity of categorization patterns to discrete types'' \citep{Walchli:2012ling}. In fact, language vectors \citep{malaviya2017learning,bjerva2018phonology} and distributional information from multi-parallel texts \citep{asgari2017past} are promising insofar they capture latent properties of languages in a bottom-up fashion, preserving their gradient nature. This offers an alternative to hand-crafted database features: in \autoref{ssec:decoding}
we make a case for integrating continuous, data-driven typological representations into NLP algorithms.

%% file: 5_use.tex
The typological features developed as discussed in \S\ \ref{sec:survey-development} are of significant importance for NLP algorithms. Particularly, they are employed in three main ways. First, they can be manually converted into rules for expert systems (\S \ref{ssec:manualrules}); second, they can be integrated into algorithms as constraints that inject prior knowledge or tie together specific parameters across languages (\S\ \ref{ssec:modelfeatures}); and, finally, they can guide data selection and synthesis (\S\ \ref{ssec:dataselection}). All of these approaches are summarized in Table \ref{tab:uses} and described in detail in the following sections, with a particular focus on the second approach.

\begin{table}[t]
\begin{center}
\def\arraystretch{1.00}
\begin{footnotesize}
\begin{tabularx}{\linewidth}{smmnm}
\toprule
 & {\bf Author} & {\bf Details} & {\bf Number of Languages / Families} & {\bf Task} \\
\midrule

\parbox[t]{2mm}{\multirow{1}{*}{Rules}} & \citet{bender2016linguistic} & Grammar generation & 12 / 8 & semantic parsing\\
\hline

\parbox[t]{2mm}{\multirow{6}{*}{\rotatebox[origin=c]{90}{Feature engineering}}} & \citet{Naseem-2012} & Generative & 17 / 10 & syntactic parsing\\
& \citet{Tackstrom-2013} & Discriminative graph-based & 16 / 7 & syntactic parsing\\
& \citet{Zhang-2015} & Discriminative tensor-based & 10 / 4 & syntactic parsing\\
\cline{2-5}
& \citet{daiber2016universal} & One-to-many MLP & 22 / 5 & reordering for machine translation\\
& \citet{Ammar-2016} & Multi-lingual transition-based & 7 / 1 & syntactic parsing\\
& \citet{Tsvetkov-2016} & Phone-based polyglot language model & 9 / 4 & identification of lexical borrowings and speech synthesis\\
\cline{2-5}
& \citet{schone2001language} & Design of Bayesian network & 1 / 1 & word cluster labeling\\

\hline

\parbox[t]{2mm}{\multirow{5}{*}{\rotatebox[origin=c]{90}{Data Manipulation}}} & \citet{Deri-2016} & Typology-based selection & 227 & grapheme to phoneme\\
& \citet{agic2017cross} & PoS divergence metric & 26 / 5 & syntactic parsing\\
& \citet{sogaard2012empirical} & Typology-based weighing & 12 / 1 & syntactic parsing\\
\cline{2-5}
& \citet{wang-eisner-2017} & Word-order-based tree synthesis & 17 / 7 & syntactic parsing\\
& \citet{ponti2018isomorphic} & Construction-based tree preprocessing & 6 / 3 & machine translation, sentence similarity\\

\bottomrule
\end{tabularx}
\end{footnotesize}
\end{center}
\vspace{-0.0em}
\caption{An overview of the approaches to use typological features in NLP models.}
\vspace{-0.1em}
\label{tab:uses}
\end{table}

\subsection{Rule-based Systems}
\label{ssec:manualrules}


An interesting example of a rule-based system in our context is the Grammar Matrix kit, presented by \citet{bender2016linguistic}, where 
rule-based grammars can be generated from typological features.
These grammars are designed within the framework of Minimal Recursion Semantics \citep{copestake2005minimal} and can parse a natural language input string into a semantic logical form. 

The Grammar Matrix consists of a universal core grammar and language-specific libraries for phenomena where typological variation is attested. For instance, the module for coordination typology expects the specification of the kind, pattern, and position of a grammatical marking, as well as the phrase types it covers. For instance, the Ono language (Trans–New Guinea) expresses it with a lexical, monosyndetic, pre-nominal marker \textit{so} in noun phrases. A collection of pre-defined grammars is available through the Language CoLLAGE initiative \citep{bender2014language}.

\subsection{Feature Engineering and Constraints}
\label{ssec:modelfeatures}

The most common usage of typological features in NLP is in feature engineering and constraint design for machine learning algorithms. Two popular approaches we consider here are \textit{language transfer with selective sharing}, where the parameters of languages with similar typological features are tied together (\S\ \ref{ssec:selshar}), and \textit{joint multilingual learning}, where typological information is used in order to bias models to reflect the properties of specific languages (see \S\ \ref{ssec:multibias}). 

\subsubsection{Selective sharing}
\label{ssec:selshar}
This framework was introduced by \citet{Naseem-2012} and was subsequently adopted by \namecite{Tackstrom-2013} and \citet{Zhang-2015}. It aims at parsing sentences in a language transfer setting (see \autoref{ssec:langtrasf}) where there are multiple source languages and a single unobserved target language. It assumes that head--modifier relations between part of speech pairs are universal, but the order of parts of speech within a sentence is language-specific. For instance, adjectives always modify  nouns, but in Igbo (Niger--Congo) they linearly precede nouns, while in Nihali (isolate) they follow nouns. Leveraging this intuition, selective sharing models learn dependency relations from all source languages, while ordering is learned from typologically related languages only.

Selective sharing was originally implemented in a generative framework, factorizing the recursive generation of dependency tree fragments into two steps \citep{Naseem-2012}. The first one is universal: the algorithm selects an unordered (possibly empty) set of modifiers $\{M\}$ given a head \textit{h} with probability $P(\{M\} | h)$, where both the head and the modifiers are characterized by their PoS tags. The second step is language-specific: each dependent \textit{m} is assigned a direction ${d}$ (left or right) with respect to \textit{h} based on the language \textit{l}, with probability $P({d}|m,h,l)$. Dependents in the same direction are eventually ordered with a probability drawn from a uniform distribution over their possible unique permutations. The total probability is then defined as follows:
\begin{equation} \label{eq:9}
P(n|h,\theta_{1}) \cdot \sigma_n\left( \sum_{m_i \in M} P(m_{i}|h,\theta_2) \right) \cdot \prod_{m_i \in M} \sigma\left( \mathbf{w} \cdot \mathbf{g}(m,h,l,\mathbf{f}_l) \right) \cdot \frac{1}{||M_R|| ||M_L||}
\end{equation}
In Equation~\ref{eq:9}, the first step is expressed as two factors: the estimation of the number \textit{n} of modifiers, parametrized by $\theta_1$, and the actual selection of modifiers,  parametrized by $\theta_2$, with the softmax function $\sigma$ converting the \textit{n} values into probabilities. The second step, overseeing the assignment of a direction to the dependencies, is parametrized by $\mathbf{w}$ that multiplies a feature function $\mathbf{g}()$, whose arguments include a typology feature vector $\mathbf{f_l}$. The values of all the parameters are estimated by maximizing the likelihood of the observations.

\citet{Tackstrom-2013} proposed a discriminative version of the model, in order to amend the alleged limitations of the original generative variant. In particular, they dispose of the strong independence assumptions (e.g.\ between choice and ordering of modifiers) and invalid feature combinations. For instance, the WALS feature `Order of subject, verb, and object' (81A) should be taken into account only when the head under consideration is a verb and the dependent is a noun, but in the generative model this feature was fed to $\mathbf{g}()$ regardless of the head--dependency pair. The method of \citet{Tackstrom-2013} is a delexicalized first-order graph-based parser based on a carefully selected feature set. From the set proposed by \citet{mcdonald2005online}, they keep only (universal) features describing selectional preferences and dependency length. Moreover, they introduce (language-specific) features for the directionality of dependents, based on combinations of the PoS tags of the head and modifiers with corresponding WALS values.

This approach was further extended to tensor-based models by \namecite{Zhang-2015}, in order to avoid the shortcomings of manual feature selection. They induce a compact hidden representation of features and languages by factorizing a tensor constructed from their combination. The prior knowledge from the typological database enables the model to forbid the invalid interactions, by generating intermediate feature embeddings in a hierarchical structure. In particular, given \textit{n} words and \textit{l} dependency relations, each arc $h \rightarrow m$ is encoded as the tensor product of three feature vectors for heads $\Phi_h \in \mathbb{R}^n$, modifiers $\Phi_m \in \mathbb{R}^n$ and the arcs $\Phi_{h\rightarrow m} \in \mathbb{R}^l$. A score is obtained through the inner product of these and the corresponding \textit{r} rank-1 dense parameter matrices for heads $H \in \mathbb{R}^{n\times r}$, dependents $M \in \mathbb{R}^{n\times r}$, and arcs $M \in \mathbb{R}^{l\times r}$. The resulting embedding is subsequently constrained through a summation with the typological features $T_u \phi_{t_u}$:
\begin{equation} \label{eq:10}
\begin{split}
S(h \xrightarrow{l} m) = \sum_{i=1}^{r} [ H_c \phi_{h_c} ]_i [ M_c \phi_{m_c} ]_i \, \odot \\
\{ [ T_l \phi_{t_l} ]_i + [ L \phi_{l} ]_i \, \odot\\ 
\left(  [ T_u \phi_{t_u} ]_i + [ H \phi_{h} ]_i [ M \phi_{m} ]_i [ D \phi_{d} ]_i \right) \}
\end{split}
\end{equation}
Equation \ref{eq:10} shows how the overall score of a labeled dependency is enriched (by element-wise product) with (1) the features and parameters for arc labels $L \phi_{l}$ constrained by the typological vector $T_l \phi_{t_l}$; and (2) features and parameters for head contexts $H_c \phi_{h_c}$ and dependent contexts $M_c \phi_{m_c} $. This loss is optimized within a maximum soft-margin objective through on-line passive--aggressive updates.


The different approaches to selective sharing presented here explicitly deal with cases where the typological features do not match any of the source languages, which may lead learning astray. 
\citet{Naseem-2012} propose a variant of their algorithm where the typological features are not observed (in WALS), treating them as latent variables, and learning the model parameters in an unsupervised fashion with the Expectation Maximization algorithm \citep{dempster1977maximum}. 
\citet{Tackstrom-2013} tackle the same problem from the side of ambiguous learning. The discriminative model on the target language is trained on sets of automatically predicted ambiguous labels $\widehat{\mathbf{y}}$. 
Finally, \citet{Zhang-2015} employ semi-supervised techniques, where only a handful of annotated examples from the target language is available.


\subsubsection{Multi-lingual Biasing}
\label{ssec:multibias} 

Some papers leverage typological features to gear the shared parameters of a joint multilingual model toward the properties of a specific language. \citet{daiber2016universal} develop a reordering algorithm that estimates the permutation probabilities of aligned word pairs in multi-lingual parallel texts. The best sequence of permutations is inferred via k-best graph search in a finite state automaton, producing a lattice. This algorithm, which receives lexical, morphological, and syntactic features of the source word pairs and typological features of the target language as input, has shown to benefit a downstream machine translation task. 


The joint multilingual parser of \citet{Ammar-2016} shares hidden-layer parameters across languages, and combines both language-invariant and language-specific features in its copious lexicalized input feature set. 
This transition-based parser selects the next action \textit{z} (e.g.\ \textsc{shift}) from a pool of possible actions given its current state $\textbf{p}_t$, as defined in Equation \ref{eq:ammar}:
\begin{equation} \label{eq:ammar}
P(z|\textbf{p}_t) = \sigma( \mathbf{g_z}^\top \text{max} (\mathbf{0}, \mathbf{W} \, \mathbf{s}_t \oplus \mathbf{b}_t \oplus  \mathbf{a}_t \oplus \mathbf{l}_{it} + \mathbf{b}) + q_z)
\end{equation}
$P(z|\textbf{p}_t)$ is defined in terms of a set of iteratively manipulated, densely represented data structures: a buffer $\textbf{b}_t$, a stack $\textbf{s}_t$, and an action history $\textbf{a}_t$. The hidden representation of these modules are the output of stack-LSTMs, that are in turn fed with input word feature representations (stack and buffer) and action representations (history). The shared parameters are biased toward a particular language through language embeddings $\mathbf{l}_{it}$. The language embeddings consist of (a non-linear transformation of) either a mere one-hot identity vector or a vector of typological properties taken from WALS. In particular, they are added to both input feature and action vectors, to affect the three above-mentioned modules individually, and concatenated to the hidden module representations, to affect the entire parser state. The resulting state representation is propagated through an action-specific layer parametrized by $\textbf{g}_t$ and $q_t$, and activated by a softmax function $\sigma$ over actions.

Similarly, typological features have been employed to bias input and hidden states of language models. For example, \citet{Tsvetkov-2016} proposed a multilingual phoneme-level language model where an input phoneme \textbf{x} and a language vector $\ell$ at time \textit{t} are linearly mapped to a local context representation and then passed to a global LSTM. This hidden representation $\textbf{G}_t^\ell$ is factored by a non-linear transformation of typological features $t_\ell$, as shown in Equation \ref{eq:tsvetkov}:
\begin{gather} \label{eq:tsvetkov}
\textbf{G}_t^\ell = \text{LSTM}(W_{c_x} x_t + W_{c_\ell} x_\ell + b, \mathbf{g}_{t-1}) \otimes \text{tanh}(W_\ell \, t_\ell + b_\ell)^\top \\
\label{eq:tsvetkov2}
P(\phi_t | \phi_{<t}, \ell) = \sigma (\textbf{W} \, \text{vec}(\textbf{G}_t^\ell) + \textbf{b})
\end{gather}
As described in Equation \ref{eq:tsvetkov2}, $\textbf{G}_t^\ell$ is then vectorized and mapped to a probability distribution of possible next phonemes $\phi_t$. The phoneme vectors, learned by the language model in an end-to-end manner, were demonstrated to benefit two downstream applications: lexical borrowing identification and speech synthesis.

Moreover, typological features (in the form of implicational universals) can guide the design of Bayesian networks. \citet{schone2001language} assign part-of-speech labels to word clusters acquired in an unsupervised fashion. The underlying network is acyclic and directed, and is converted to a join-tree network to handle multiple parents \citep{jensen1996introduction}. For instance, the sub-graph for the ordering of numerals and nouns is intertwined also with properties of adjectives and adpositions. The final objective maximizes the probability of a tag $T_i$ and a feature set $\Phi_i$ given the implicational universals $U$ as $\text{argmax}_{T} P(\{\Phi_i, T_i\}_{i=1}^n | U)$.

\subsection{Data Selection, Synthesis, and Preprocessing}
\label{ssec:dataselection}
Another way in which typological features are used in NLP is to guide data selection. This procedure is crucial for 1) language transfer methods, as it guides the choice of the most suitable source languages and examples; and 2) multilingual joint models, in order to weigh the contribution of each language and example. The selection is typically carried out through general language similarity metrics. 
 For instance, \citet{Deri-2016} base their selection on the URIEL language typology database, considering information about genealogical, geographic, syntactic, and phonetic properties. 
This facilitates language transfer of  grapheme-to-phoneme models, by guiding the choice of source languages and aligning phoneme inventories. 

Metrics for source selection can also be extracted in a data-driven fashion, without explicit reference to structured taxonomies. For instance, \citet{Rosa:2015acl} estimate the Kullback--Leibler divergence between part-of-speech trigram distributions for delexicalized parser transfer. In order to approximate the divergence in syntactic structures between languages, \namecite{ponti2018isomorphic} employ the Jaccard distance between morphological feature sets and the tree edit distance of delexicalized dependency parses of similar sentences.

A-priori and bottom-up approaches can also be combined. For delexicalized parser transfer, \citet{agic2017cross} relies on a weighted sum of distances based on 1) the PoS divergence defined by \citet{Rosa:2015acl}; 2) the character-based identity prediction of the target language; and 3) the Hamming distance from the target language typological vector. In fact, they have different weaknesses: language identity (and consequently typology) fail to abstract away from language scripts. On the other hand, the accuracy of PoS-based metrics deteriorates easily in scenarios with poor amounts of data.

Source language selection is a special case of source language weighting where weights are one-hot vectors. However, weights can also be gradient and consist of real numbers. \citet{sogaard2012empirical} adapt delexicalized parsers by weighting every training instance based on the inverse of the Hamming distance between typological (or genealogical) features in source and target languages. An equivalent bottom-up approach is developed by \namecite{sogaard-2011} who weighs source language sentences based on the perplexity between their coarse PoS tags and the predictions of a sequential model trained on the target language.

Alternatively, the lack of target annotated data can be alleviated by synthesizing new examples, thus boosting the variety and amount of the source data. For instance, the Galactic Dependency Treebanks stem from real trees whose nodes have been permuted probabilistically according to the word order rules for nouns and verbs in other languages \cite{wang2016galactic}. 
Synthetic trees improve the performance of model transfer for parsing when the source is chosen in a supervised way (performance on target development data) and in an unsupervised way (coverage of target PoS sequences). 

Rather than generating new synthetic data,
\citet{ponti2018isomorphic} leverage typological features to pre-process treebanks in order to reduce their variation in language transfer tasks. In particular, they adapt source trees to the typology of a target language with respect to several constructions in a rule-based fashion. 
For instance, relative clauses in Arabic (Afro--Asiatic) with an indefinite antecedent drop the relative pronoun, which is mandatory in Portuguese (Indo--European). Hence, the pronoun has to be added, or deleted in the other direction.
Feeding pre-processed syntactic trees to lexicalized syntax-based neural models, such as feature-based recurrent encoders \citep{sennrich2016linguistic} or TreeLSTMs \citep{tai2015improved}, achieves state-of-the-art results in Neural Machine Translation and cross-lingual sentence similarity classification. 

\begin{figure}[t!]
    \centering
    \includegraphics[width=\textwidth]{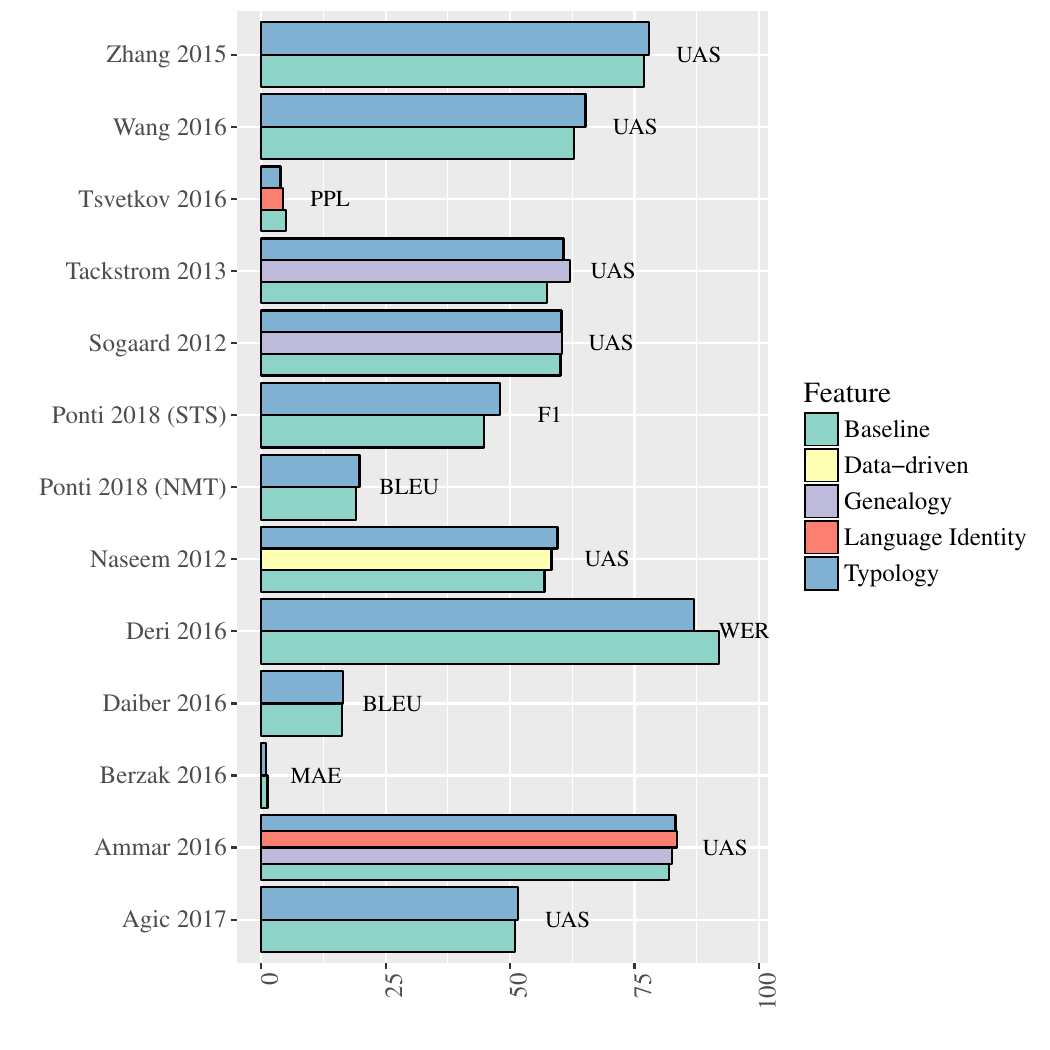}
    \caption{Performance of the surveyed algorithms for the tasks detailed in Table \ref{tab:uses}. The algorithms are evaluated with different feature sets: no typological features (Baseline), latently inferred typology (Data-driven), Genealogy, Language Identity, and gold database features (Typology). Evaluation metrics are reported right of the bars: Unlabeled Attachment Score (UAS), Perplexity (PPL), F1 Score, BiLingual Evaluation Understudy (BLEU), Word Error Rate (WER), and Mean Absolute Error (MAE).}
    \label{fig:performance_all}
\end{figure}


\subsection{Comparison}
\label{ssec:comptask}
In light of the performance of the above methods, to what extent can typological features benefit downstream NLP tasks and applications? To answer this key question, consider the performance scores of each model reported in Figure \ref{fig:performance_all}. Each model has been evaluated in the original paper in one (or more) of the three main settings: with gold database features (Typology), with latently inferred typological features (Data-driven), or without both (Baseline), and with otherwise identical architecture and hyper-parameters. 

It is evident that typology-enriched models consistently outperform baselines across several NLP tasks. Indeed, the scores are higher for metrics that increase (Unlabeled Attachment Score, F1 Score and BLEU) and lower for metrics that decrease (Word Error Rate, Mean Average Error and Perplexity) with better predictions. 
Nevertheless, improvements tend to be moderate, and only a small number of experiments support them with statistical significance tests. In general, it appears that they fall short of the potential usefulness of typology: in \autoref{sec:future} we analyse the possible reasons for this.


Some of the experiments we have surveyed investigate the effect of substituting typological features with features related to Genealogy and Language Identity (e.g.\ one-hot encoding of languages). Based on the results in Figure \ref{fig:performance_all}, it is unclear whether typology should be preferred, as it is sometimes rivaled by other types of features. In particular, it is typology that excels according to \citet{Tsvetkov-2016}, genealogy according to \citet{sogaard2012empirical} and \citet{Tackstrom-2013}, and language identity according to \citet{Ammar-2016}. However, drawing conclusions from the last experiment seems incautious: in \autoref{sec:manual-extraction}, we argued that their selection of features (presented in Figure \ref{fig:lanreps}) is debatable due to low diversification or noise. Moreover, it should be emphasized that one-hot language encoding is limited to the joint multilingual learning setting: since it does not convey any information, it is of no avail in language transfer.



Finally, let us consider the effectiveness of the above methods with respect to incorporating typological features in NLP models. In case of selective sharing, the tensor-based discriminative model \citep{Zhang-2015} outperforms the graph-based discriminative model \citep{Tackstrom-2013}, which in turn surpasses the generative model \citep{Naseem-2012}. With regard to biasing multilingual models, there is a clear tendency toward letting typological features interact not merely with the input representation, but also with deeper levels of abstraction such as hidden layers.

Overall, this comparison supports the claim that typology can potentially aid in designing the architecture of algorithms, engineering their features, and selecting and pre-processing their data. Nonetheless, this discussion also revealed that many challenges lie ahead for each of these goals to be accomplished fully. We discuss them in the next section.


%% file: 6_future.tex
In \autoref{sec:survey-applications} we surveyed the current uses of typological information in NLP. In this section we discuss potential future research avenues that may result in a closer and more effective integration of linguistic typology and multilingual NLP. In particular, we discuss: 1) the extension of existing methods to new tasks, possibly exploiting typological resources that have been neglected thus far (\S\ \ref{ssec:newtasks}); 2) new methods for injecting typological information into NLP models as soft constraints or auxiliary objectives (\S\ \ref{ssec:encoding}); and 3) new ways to acquire and represent typological information that reflect the gradient and contextual nature of cross-lingual variation  (\S\ \ref{ssec:decoding}).

\subsection{Extending the Usage to New Tasks and Features}
\label{ssec:newtasks}

The trends observed in \autoref{sec:survey-applications} reveal that typology is integrated into NLP models mostly in the context of morphosyntactic tasks, and particularly syntactic parsing. Some exceptions include other levels of linguistic structure, such as phonology \citep{Tsvetkov-2016,Deri-2016} and semantics \citep{bender2016linguistic,ponti2018isomorphic}. As a consequence, the set of selected typological features is mostly limited to a handful of word-order features from a single database, WALS. Nonetheless, the array of tasks that pertain to polyglot NLP is broad, and other typological datasets that have thus far been neglected may be relevant for them.

For example, typological frame semantics might benefit semantic role labeling, as it specifies the valency patterns of predicates across languages, including the number of arguments, their morphological markers, and their order. This information can be cast in the form of priors for unsupervised syntax-based Bayesian models \citep{titov2012crosslingual}, guidance for alignments in annotation projection \citep{pado2009cross,van2011scaling}, or regularizers for model transfer in order to tailor the source model to the grammar of the target language \citep{kozhevnikov2013cross}. Cross-lingual information about frame semantics can be extracted, for example, from the Valency Patterns Leipzig database (ValPaL).

Typological information regarding lexical semantics patterns can further assist various NLP tasks, by providing information about translationally equivalent words across languages. Such information is provided in databases such as the World Loanword Database (WOLD), the Intercontinental Dictionary Series (IDS), and the Automated Similarity Judgment Program (ASJP). One example task is word sense disambiguation, as senses can be propagated from multilingual word graphs \citep{silberer2010uhd} by bootstrapping from a few pivot pairs \cite{Khapra-2011}, by imposing constraints in sentence alignments and harvesting bag-of-words features from these \cite{lefever2011parasense}, or by providing seeds for multilingual WE-based lexicalized model transfer \citep{zennaki2016inducing}. 

Another task where lexical semantics is crucial is sentiment analysis, for similar reasons: bilingual lexicons constrain word alignments for annotation projection \citep{almeida2015aligning} and provide pivots for shared multilingual representations in model transfer \citep{fernandez2015distributional,Ziser:18}. Moreover, sentiment analysis can leverage morphosyntactic typological information about constructions that alter polarity, such as negation \citep{ponti2017decoding}.

Finally, morphological information was shown to aid interpreting the intrinsic difficulty of texts for language modeling and neural machine translation, both in supervised \cite{johnson2016google} and in unsupervised \citep{artetxe2017unsupervised} setups. In fact, the degree of fusion between roots and inflectional/derivative morphemes impacts the type/token ratio of texts, and consequently their rate of infrequent words. Moreover, the ambiguity of mapping between form and meaning of morphemes determines the usefulness of injecting character-level information \citep{gerz2018language,gerz-etal-2018-relation}. This variation has to be taken into account in both language transfer and multilingual joint learning.

As a final note, we stress that the addition of new features does not concern just future work, but also the existing typology-savvy methods, which can widen their scope. For instance, the parsing experiments grounded on selective sharing (\autoref{ssec:modelfeatures}) could also take into consideration WALS features about Nominal Categories, Nominal Syntax, Verbal Categories, Simple Clauses, and Complex Sentences, as well as features from other databases such as SSWL, APiCS, and AUTOTYP. Likewise, models for phonological tasks \citep{Tsvetkov-2016,Deri-2016} could also extract features from typological databases such as LAPSyD and StressTyp2.


\subsection{Injecting Typological Information into Machine Learning Algorithms}
\label{ssec:encoding}

In \autoref{sec:survey-applications}, we discussed the potential of typological information to provide guidance to NLP methods, and surveyed approaches such as network design in Bayesian models \citep{schone2001language}, selective sharing \citep[\textit{inter alia}]{Naseem-2012}, and biasing of multilingual joint models \citep[\textit{inter alia}]{Ammar-2016}. However, many other frameworks (including those already mentioned in \autoref{sec:applications}) have been developed independently in order to allow the integration of expert and domain knowledge into traditional feature-based machine learning algorithms and neural networks. In this section we survey these frameworks and discuss their applicability to the integration of typological information into NLP models.  

Encoding cross-language variation and preferences into a machine learning model requires a mechanism that can bias the \textit{learning} (i.e. training and parameter estimation) and \textit{inference} (prediction) of the model towards some pre-defined knowledge. In practice, learning algorithms, both linear (e.g. structured perceptron \cite{Collins:02}, MIRA \cite{Crammer:03} and structured SVM \cite{Taskar:04}) and non-linear (deep neural models) iterate between an inference step and a step of parameter update with respect to a gold standard. The inference step is the natural place where external knowledge could be encoded through constraints. This step biases the prediction of the model to agree with the external knowledge which, in turn, affects both the training process and the final prediction of the model at test time.

Information about cross-lingual variation, particularly when extracted empirically (see \autoref{sec:survey-development}), reflects tendencies rather than strict rules. As a consequence, {\it soft}, rather than {\it hard constraints} are a natural vehicle for their encoding. We next survey a number of existing approaches that can efficiently encode such constraints. 

The goal of an inference algorithm is to predict the best output label according to the current state of the model parameters.\footnote{Generally speaking, an inference algorithm can make other predictions such as computing expectations and marginal probabilities. As in the context of this paper we are mostly focused on the prediction of the best output label, we refer only to this type of inference problems.} For this purpose, the algorithm searches the space of possible output labels in order to find the best one. Efficiency hence plays a key role in these algorithms. Introducing soft constraints into an inference algorithm, therefore, posits an algorithmic challenge: how can the output of the model be biased to agree with the constraints while the efficiency of the search procedure is kept? In this paper we do not answer this question directly but rather survey a number of approaches that succeed in dealing with it.

Since linear models have been prominent in NLP research for a much longer time, it is not surprising that frameworks for the integration of soft constraints into these models are much more developed. The approaches proposed for this purpose include  posterior regularization (PR) \cite{Ganchev:10}, generalized expectation (GE) \cite{Mann:08}, constraint-driven learning (CODL) \cite{Chang:07}, dual decomposition (DD) \cite{Globerson:08,Komodakis:11} and Bayesian modeling \cite{cohen-16}. These techniques employ different types of knowledge encoding, e.g. PR uses expectation constraints on the posterior parameter distribution, GE prefers parameter settings where the model’s distribution on unsupervised data matches a predefined target distribution, CODL enriches existing statistical models with Integer Linear Programming (ILP) constraints, while in Bayesian modeling a prior distribution is defined on the model parameters.

PR has already been used for incorporating universal linguistic knowledge into an unsupervised parsing model \cite{Naseem:10}. In the future, it could be extended to typological knowledge, which is a good fit for soft constraints. As another option, Bayesian modeling sets prior probability distributions according to the relationships encoded in typological features \citep{schone2001language}. Finally, DD has been applied to multi-task learning, which paves the way for typological knowledge encoding through a multi-task architecture in which one of the tasks is the actual NLP application and the other is the data-driven prediction of typological features. In fact a modification of this archiecture has already been applied to minimally supervised learning and domain adaptation with soft (non-typological) constraints~\citep{Rush:12,Reichart-2012b}.

The same ideas could be exploited in deep learning algorithms. We have seen in \S\ \ref{ssec:jointlearn} that multilingual joint models combine both shared and language-dependent parameters, in order to capture the universal properties and cross-lingual differences, respectively. In order to enforce this division of roles more efficiently, these models could be augmented with the auxiliary task of predicting typological features automatically. This auxiliary objective could update parameters of the language-specific component, or those of the shared component, in an adversarial fashion, similarly to what \citet{chen2016adversarial} implemented by predicting language identity. 

Recently, \citet{Hu:16a,Hu:16b} and \citet{Wang:18} proposed frameworks that integrate deep neural models with manually specified or automatically induced constraints. Similarly to CoDL, the focus in \citet{Hu:16a} and \citet{Wang:18} is on logical rules, while the ideas in \citet{Hu:16b} are related to PR. These frameworks provide a promising avenue for the integration of typological information and deep models.



A particular non-linear deep learning domain where knowledge integration is already prominent is multilingual representation learning (\S\ \ref{subsec:reprlearn}). In this domain a number of works \cite{Faruqui:2015naacl,Rothe:2015acl,osborne-16,mrksic:2016:naacl} have proposed means through which external knowledge sourced from linguistic resources (such as WordNet, BabelNet, or lists of morphemes) can be encoded in word embeddings. Among the state-of-the-art specialization methods \textsc{attract-repel} \citep{mrkvsic2017semantic,vulic2017morph} pushes together or pulls apart vector pairs according to relational constraints, while preserving the relationship between words in the original space and possibly propagating the specialization knowledge to unseen words or transferring it to other languages \citep{ponti2018adversarial}. The success of these works suggests that a more extensive integration of external linguistic knowledge in general, and typological knowledge in particular, is likely to play a key role in the future development of word representations.

\subsection{A New Typology: Gradience and Context-Sensitivity}
\label{ssec:decoding}

As shown in \autoref{sec:manual-extraction}, most of the typology-savvy algorithms thus far exploited features extracted from manually crafted databases. However, this approach is riddled by several shortcomings, which are reflected in the small performance improvements observed in \autoref{ssec:comptask}. Luckily, these shortcomings may potentially be averted through the use of methods 
 that allow typological information to emerge from the data in a bottom-up fashion, rather than being predetermined. In what follows we advocate for such a data-driven approach, based on several considerations.

Firstly, typological databases provide \textit{incomplete} documentation of the cross-lingual variation, in terms of features and languages. Raw textual data, which is easily accessible for many languages and cost-effective, may provide a valid alternative that can facilitate automatic learning of more complete knowledge. Secondly,  database information is \textit{approximate}, as it is restricted to the majority strategy within a language. However, in theory each language allows for multiple strategies in different \textit{contexts} and with different \textit{frequencies}, hence databases risk hindering models from learning less likely but plausible patterns \citep{sproat2016language}. Inferring typological information from text would enable a system to discover patterns within individual examples, including both the frequent and the infrequent ones. Thirdly, typological features in databases are \textit{discrete}, employing predefined categories devised to make high-level generalizations across languages. However, several categories in natural language are \textit{gradient} (see for instance the discussion on semantic categorization in \autoref{sec:overview}), hence they are better captured by continuous features. In addition to being psychologically motivated, this sort of gradient representation is also more compatible with machine learning algorithms and particularly with deep neural models that naturally operate with real-valued multi-dimensional word embeddings and hidden states.

To sum up, the automatic development of typological information and its possible integration into machine learning algorithms have the potential to solve an important bottleneck in polyglot NLP. Current manually curated databases consist of incomplete, approximate, and discrete features that are intended to reflect contextual and gradient information implicitly present in text. These features are fed to continuous, probabilistic, and contextual machine learning models --- which do not form a natural fit for the typological features. Instead, we believe that modeling cross-lingual variation directly from textual data can yield typological information that is more suitable for machine learning.

Several techniques surveyed in \autoref{automatic-learning} are suited to serve this purpose. In particular, the extraction from morphosyntactic annotation \citep[\textit{inter alia}]{liu2010dependency} and alignments from multi-parallel texts \citep[\textit{inter alia}]{asgari2017past} provide information about typological constructions at the level of individual examples. Moreover, language vectors \citep{malaviya2017learning,bjerva2018phonology} and alignments from multi-parallel texts preserve the gradient nature of typology through continuous representations.

The successful integration of these components would affect the way multilingual feature engineering is performed. As opposed to using binary vectors of typological features, the information about language-internal variation could be encoded as real-valued vectors where each dimension is a possible strategy for a given construction and its relative frequency within a language.
As an alternative, selective sharing and multilingual biasing could be performed at the level of individual examples rather than languages as a whole. In particular, model parameters could be transferred among similar examples and input/hidden representations could be conditioned on contextual typological patterns. Finally, focusing on the various instantiations of a particular type rather than considering languages as indissoluble blocks would enhance data selection, similarly to what \citet{sogaard-2011} achieved using PoS n-grams for similarity measurement. The selection of similar sentences rather than similar languages as source data in language transfer is likely to yield large improvements, as demonstrated by \citet{agic2017cross} for parsing in an oracle setting.

Finally, the bottom-up development of typological features may address also radically resource-less languages that lack even raw textual data in a digital format. For this group, which still constitutes a large portion of the world's languages, there are often available reference grammars written by field linguists, which are the ultimate source for typological databases. These grammars could be queried automatically, and fine-grained typological information could be harvested through information extraction techniques. 

%% file: 7_conclusions.tex
In this article, we surveyed a wide range of approaches integrating typological information, derived from the empirical and systematic comparison of the world's languages, and NLP algorithms. The most fundamental problem for the advancement of this line of research is bridging the gap between the interpretable, language-wide, and discrete features of linguistic typology found in database documentation, and the opaque, contextual, and probabilistic models of NLP. We addressed this problem by exploring a series of questions: i) for which tasks and applications is typology useful? ii) Which are the advantages and limitations of currently available typological databases? Can data-driven inference of typological features offer an alternative source of information? iii) Which methods allow us to inject typological information from external resources, and how should such information be encoded? iv) By which margin do typology-savvy methods surpass typology-agnostic baselines? How does typology compare to other criteria of language classification, such as genealogy? v) In addition to augmenting machine learning algorithms, which other purposes do typology serve for NLP? We summarize our key findings below:

\begin{itemize}
\justifying

\item[1.] Typological information is currently used predominantly for morphosyntactic tasks, in particular dependency parsing. As a consequence, these approaches typically select a limited subset of features from a single dataset (WALS) and focus on a single aspect of variation (typically word order). However, typological databases also cover other important features, related to predicate--argument structure (ValPaL), phonology (LAPSyD, PHOIBLE, StressTyp2) and lexical semantics (IDS, ASJP), which are currently largely neglected by the multilingual NLP community. In fact, these features have the potential to benefit many tasks addressed by language transfer or joint multilingual learning techniques, such as semantic role labeling, word sense disambiguation, or sentiment analysis.

\item[2.] Typological databases tend to be incomplete, containing missing values for individual languages or features. This hinders the integration of the information in such databases into NLP models; and therefore, several techniques have been developed to predict missing values automatically. They include heuristics derived from morphosyntactic annotation; propagation from other languages based on hierarchical clusters or similarity metrics; supervised models; and distributional methods applied to multi-parallel texts. However, none of these techniques surpasses the others across the board in prediction accuracy, as each excels in different feature types. A challenge left for future work is creating ensembles of techniques to offset their individual disadvantages.

\item[3.] The most widespread approach to exploit typological features in NLP algorithms is ``selective sharing'' for language transfer. Its intuition is that a model should learn universal properties from all examples, but language-specific information only from examples with similar typological properties. Another successful approach is gearing multilingual joint models towards specific languages by concatenating typological features in input, or conditioning hidden layers and global sequence representations on them. New approaches could be inspired by traditional techniques for encoding external knowledge into machine learning algorithms through soft constraints on the inference step, semi-supervised prototype-driven methods, specialization of semantic spaces, or auxiliary objectives in a multi-task learning setting.

\item[4.] The integration of typological features into NLP models yields consistent (even if often moderate) improvements over baselines lacking such features. Moreover, guidance from typology should be preferred to features related to genealogy or other language properties. Models enriched with the latter features occasionally perform equally well due to their correlation with typological features, but fall short when it comes to modeling diversified language samples or fine-grained differences among languages.

\item[5.] In addition to feature engineering, typological information has served several other purposes. Firstly, it allows experts to define rule-based models, or to assign priors and independence assumptions in Bayesian graphical models. Secondly, it facilitates data selection and weighting, at the level of both languages and individual examples. Annotated data can be also synthesized or preprocessed according to typological criteria, in order to increase their coverage of phenomena or availability for further languages. Thirdly, typology enables researchers to interpret and reasonably foresee the difference in performance of algorithms across the sampled languages.

\end{itemize}

\noindent
Finally, we advocated for a new approach to linguistic typology inspired by the most recent trends in the discipline and aimed at averting some fundamental limitations of the current approach. In fact, typological database documentation is incomplete, approximate, and discrete. As a consequence, it does not fit well with the gradient and contextual models of machine learning. However, typological databases are originally created from raw linguistic data. An alternative approach could involve learning typology from such data automatically (i.e. from scratch). This would capture the variation within languages at the level of individual examples, and to naturally encode typological information into continuous representations. These goals have already been partly achieved by methods involving language vectors, heuristics derived from morphosyntactic annotation, or distributional information from multi-parallel texts. The main future challenge is the integration of these methods into machine learning models, as opposed to sourcing typological features from databases.


In general, we demonstrated that typology is relevant to a wide range of NLP tasks and provides the most effective and principled way to carry out language transfer and multilingual joint learning. 
 We hope that the research described in this survey will provide a platform for deeper integration of typological information and NLP techniques, thus furthering the advancement of multilingual NLP.